\documentclass[12pt]{report}

\usepackage[a4paper, margin=1in]{geometry}
\usepackage{setspace}
\usepackage{graphicx}
\usepackage{hyperref}
\usepackage{float}
\usepackage[numbers]{natbib}
\usepackage{amsmath, amssymb, amsthm, mathtools}
\usepackage{subcaption}
\usepackage{algorithm}
\usepackage{algpseudocode}
\usepackage{booktabs}
\usepackage{caption}
\usepackage{listings}
\usepackage{color}
\usepackage{microtype}

\definecolor{codegreen}{rgb}{0,0.6,0}
\definecolor{codegray}{rgb}{0.5,0.5,0.5}
\definecolor{codepurple}{rgb}{0.58,0,0.82}
\definecolor{backcolour}{rgb}{0.95,0.95,0.92}

\lstdefinestyle{mystyle}{
    backgroundcolor=\color{backcolour},   
    commentstyle=\color{codegreen},
    keywordstyle=\color{magenta},
    numberstyle=\tiny\color{codegray},
    stringstyle=\color{codepurple},
    basicstyle=\footnotesize,
    breakatwhitespace=false,         
    breaklines=true,                 
    captionpos=b,                    
    keepspaces=true,                 
    numbers=left,                    
    numbersep=5pt,                  
    showspaces=false,                
    showstringspaces=false,
    showtabs=false,                  
    tabsize=2
}

\usepackage{dsfont}
\usepackage{soul}
\usepackage{enumitem}

\begin{document}

% Title Page
\begin{titlepage}
    \centering
    {\Large\bfseries Towards General Purpose Robots at Scale:\\Lifelong Learning and Learning to Use Memory \par}
    \vspace{2cm}
    {\large Submitted in Partial Fulfillment of the Requirements for the Degree of\par}
    \vspace{1cm}
    {\large Bachelor of Science in Computer Science\\
    Turing Scholars Honors\par}
    \vspace{1cm}
    {\large by\par}
    \vspace{1cm}
    {\large William H Yue\par}
    \vfill
    {\large Committee in charge:\par}
    {\large Professor Peter Stone, Chair\\
    Professor Yuke Zhu\\
    Professor Gordon Novak\par}
    \vfill
    {\large The University of Texas at Austin\par}
    {\large Computer Science\par}
    {\large \today\par}
\end{titlepage}

% Abstract
\chapter*{Abstract}
\addcontentsline{toc}{chapter}{Abstract} % Add Abstract to TOC
The widespread success of artificial intelligence in fields like natural language processing and computer vision has not yet fully transferred to robotics, where progress is hindered by the lack of large-scale training data and the complexity of real-world tasks. To address this, many robot learning researchers are pushing to get robots deployed at scale in everyday unstructured environments like our homes to initiate a data flywheel. While current robot learning systems are effective for certain short-horizon tasks, they are not designed to autonomously operate over long time horizons in unstructured environments. This thesis focuses on addressing two key challenges for robots operating over long time horizons: \textbf{memory} and \textbf{lifelong learning}.

We propose two novel methods to advance these capabilities. First, we introduce t-DGR, a trajectory-based deep generative replay method that achieves state-of-the-art performance on Continual World benchmarks, advancing lifelong learning. Second, we develop a framework that leverages human demonstrations to teach agents effective memory utilization, improving learning efficiency and success rates on Memory Gym tasks. Finally, we discuss future directions for achieving the lifelong learning and memory capabilities necessary for robots to function at scale in real-world settings.

% Acknowledgements
\chapter*{Acknowledgements}
\addcontentsline{toc}{chapter}{Acknowledgements} % Add Acknowledgements to TOC
I would like to express my gratitude to Professor Peter Stone and Bo Liu for their invaluable guidance and support during my undergraduate studies, as well as for granting me the freedom to explore my own research interests. I am especially grateful to Bo Liu for continuously encouraging me to become a better researcher and for helping to shape my research direction throughout our time together.

% Table of Contents
\tableofcontents
\newpage

% Chapters
\chapter{Introduction}

\begin{figure}[H]
    \centering
    \includegraphics[width=\textwidth]{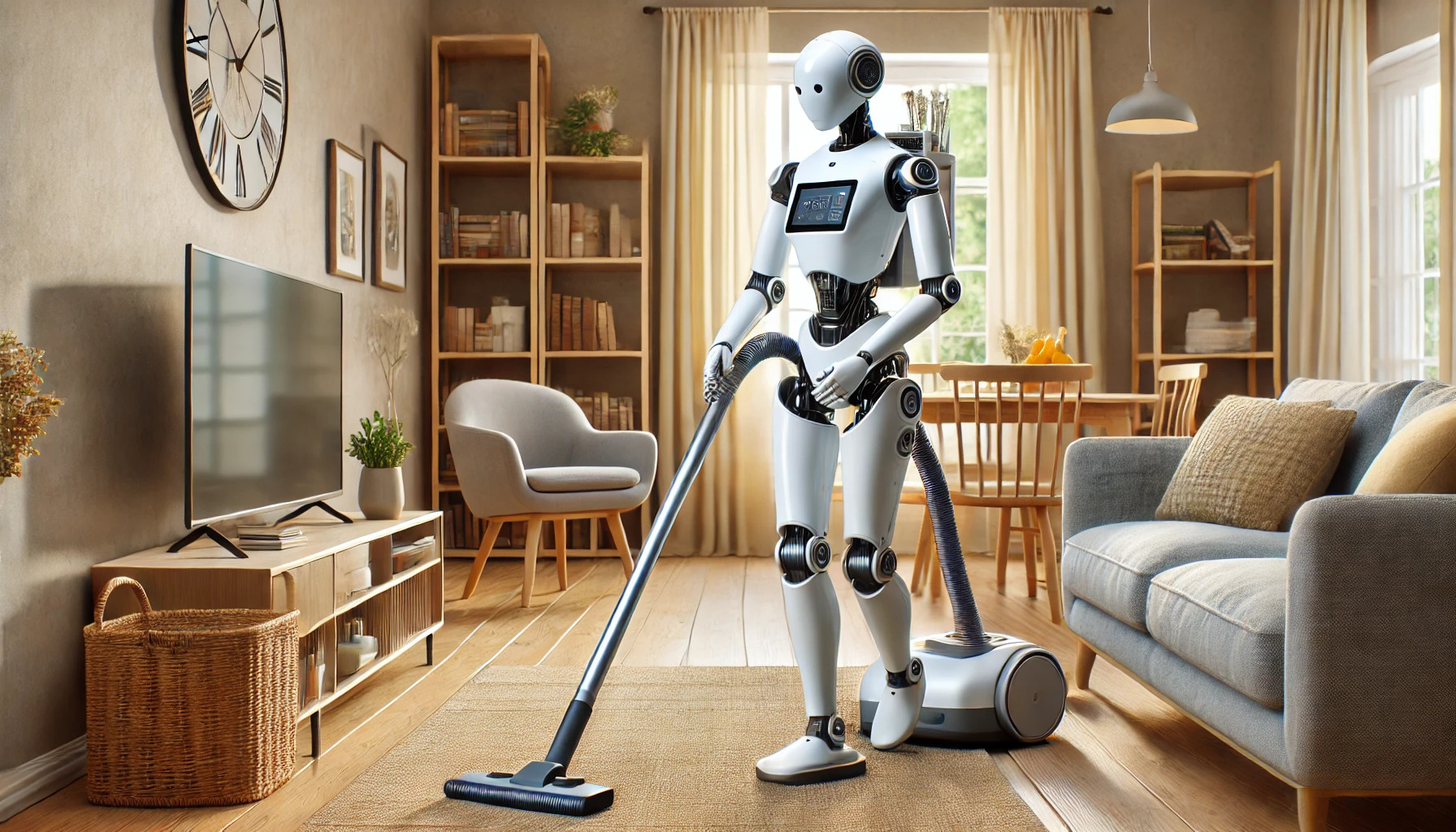}
    \caption{A humanoid operating in a living room.}
    \label{fig:humanoid}
\end{figure}

Fields like language and vision have achieved remarkable progress in advancing artificial intelligence, largely through training massive neural networks on extensive and diverse datasets \cite{hoffmann2022training, kaplan2020scaling, brown2020language, radford2021learning}. However, these successes have not fully translated to robotics, where robots are often limited to impressive demos but in controlled environments. This contrasts sharply with the widespread deployment seen in natural language processing and computer vision \cite{achiam2023gpt, betker2023improving}. A significant bottleneck in robotics progress is the lack of large-scale training data. While language and vision data are abundant online, robot data does not naturally exist at such scale \cite{khazatsky2024droid, o2023open}.

\begin{figure}[H]
    \centering
    \includegraphics[width=0.7\textwidth]{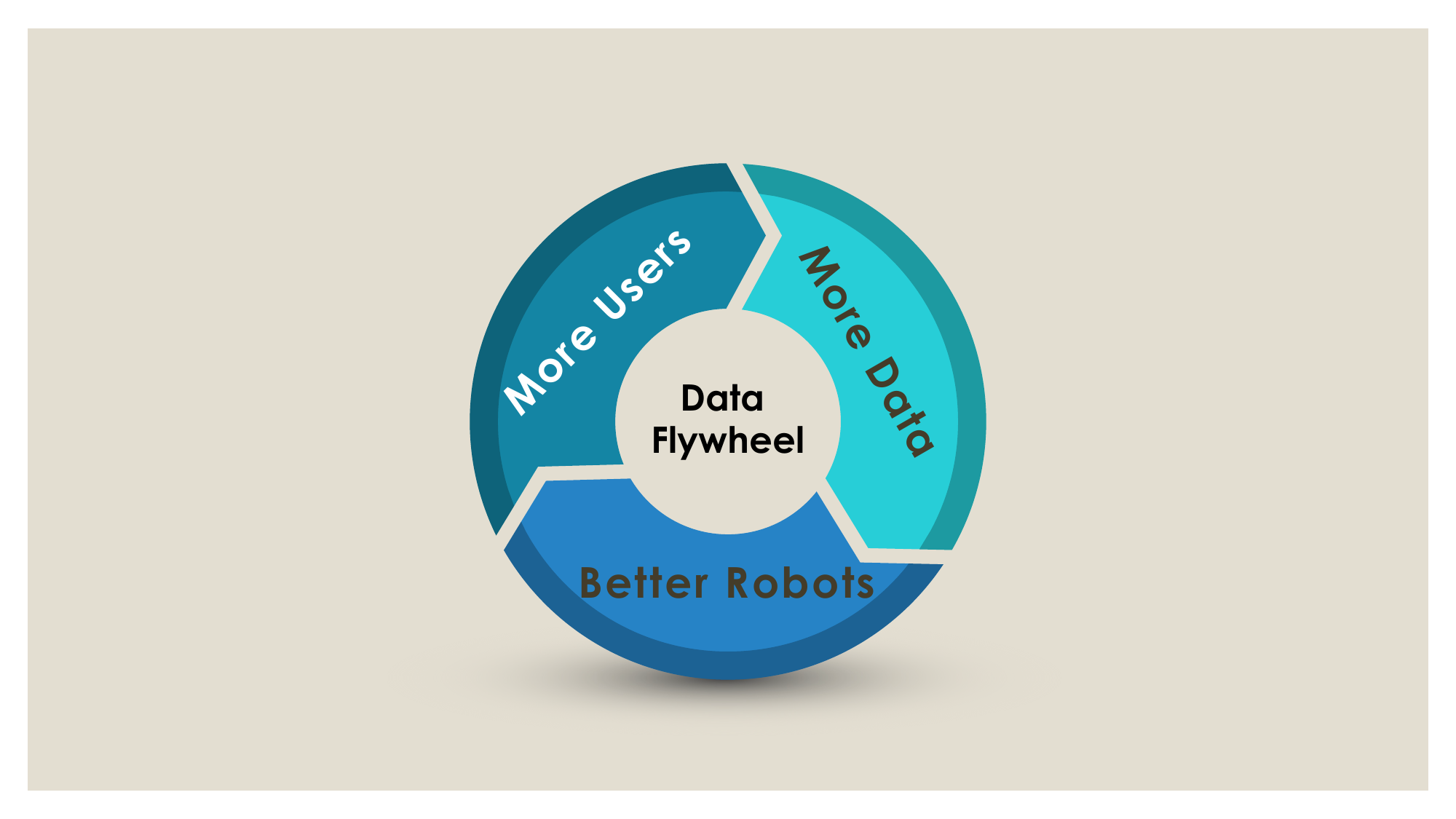}
    \caption{The data flywheel for robotics.}
    \label{fig:flywheel}
\end{figure}

A long-standing goal for robot learning researchers is to ignite a data flywheel (Figure~\ref{fig:flywheel}) through large-scale deployment of general-purpose robots \cite{levine2024generalist, zhu2023pathway}. With more users, robots would collect more diverse data, enabling better models and more capable robots, which would attract even more users \cite{collins2019turning, fridman2022karpathy}. But what does it take to achieve this data engine? To get general-purpose robots into our homes and workplaces, they must operate effectively over long-horizon timescales, far beyond the short-horizon tasks featured in today's demos. State-of-the-art robot learning algorithms are not yet designed to handle the complexity and long time-horizons required for real-world, large-scale deployment.

\begin{figure}[H]
    \centering
    \includegraphics[width=0.7\textwidth]{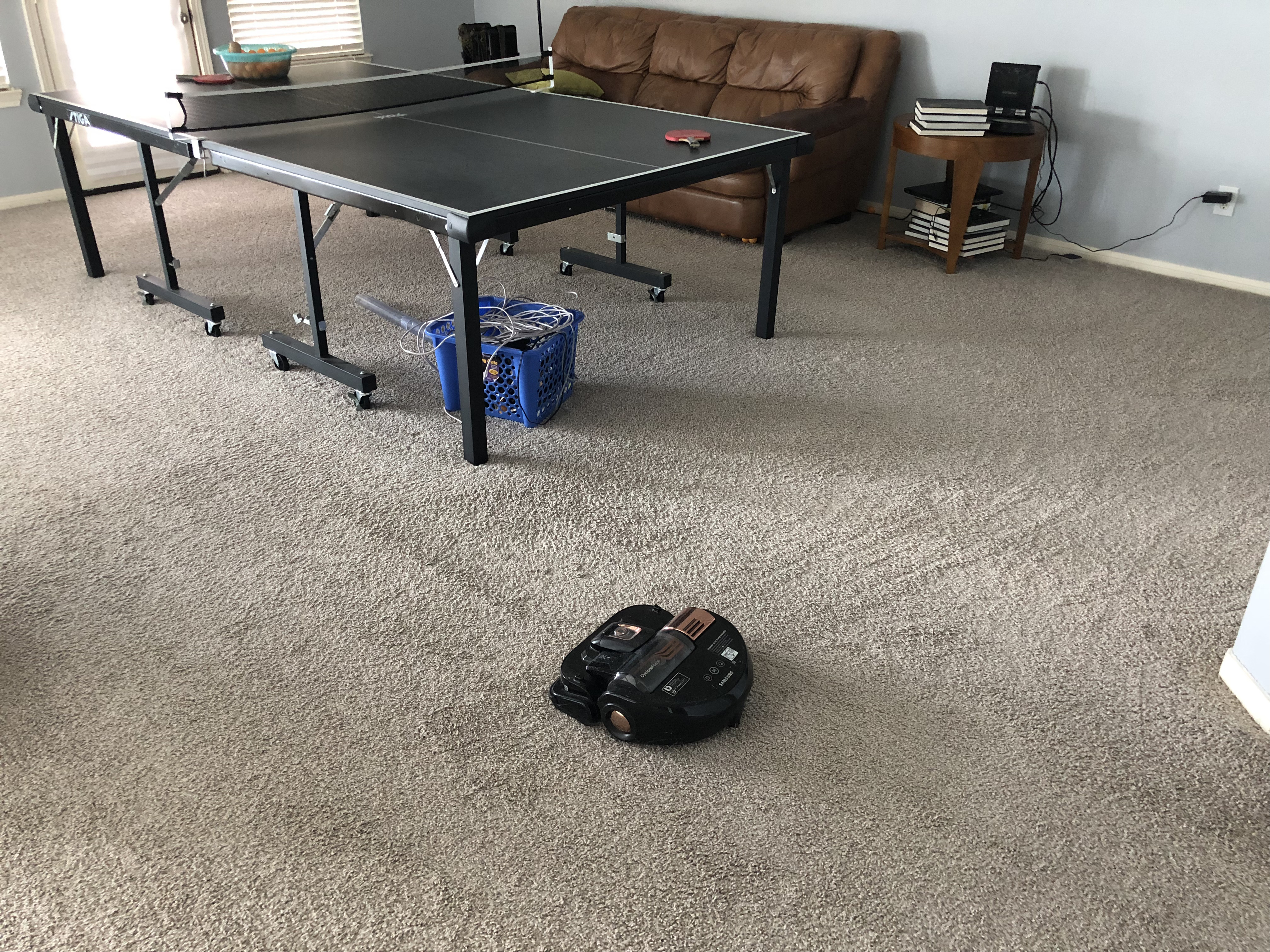}
    \caption{Roomba robot vacuuming the carpet of a residential game room.}
    \label{fig:roomba}
\end{figure}

Take the example of a Roomba vacuum robot (Figure~\ref{fig:roomba}). It may perform well in a controlled demo, cleaning a small, obstacle-free room. But in a real-world home, it requires constant user intervention—moving obstacles that get it stuck, redirecting it to missed areas—making it hardly more efficient than simply vacuuming yourself. People want robots that don’t need such babysitting. For instance, rather than setting up a robot in front of a washing machine with pre-set laundry and controls, we want robots that autonomously locate the laundry basket in a closet, carry it to the laundry room, operate the washing machine, and return clothes to their proper place.

This demands robots capable of long-horizon operation, on the scale of hours, days, or even weeks, rather than just minutes. Long-horizon functionality requires sophisticated memory to recall tasks, locations, and routines, as well as the ability to continually learn. For example, memory is crucial to remember where clothes are stored, navigate routes, and recall how items are organized. However, current methods typically rely only on immediate observations or those from a few seconds earlier, neglecting the broader context \cite{black2024pi_0, chi2023diffusionpolicy, zhao2023learning, fu2024humanplus}. Additionally, as environments evolve, robots must acquire new skills—such as adapting to changes in household layouts or factory workflows. Unfortunately, most existing approaches involve fixed models that cannot adapt post-deployment \cite{black2024pi_0}.

Crucially, achieving the memory and lifelong learning that enables long-horizon operation requires more than scaling up current methods. We need fundamentally new approaches that enable robots to develop robust memory and continuously learn throughout their deployment, addressing the complexities of real-world tasks over extended timescales. This thesis presents my efforts over the past two years to advance adaptive lifelong learning algorithms and develop capable memory mechanisms for robotic agents.

In Chapter~\ref{chp:lifelong}, we present t-DGR, a novel lifelong learning algorithm that achieves state-of-the-art performance on the Continual World benchmark. Chapter~\ref{chp:memory} introduces a framework for teaching agents to effectively utilize memory through human demonstrations. These chapters are primarily based on my papers \citep{yue2024t} and \citep{yue2024learningmemorymechanismsdecision}, respectively. Finally, Chapter~\ref{chp:conclusion} outlines potential future directions inspired by this work.

\chapter{Lifelong Learning}
\label{chp:lifelong}

In this chapter, we introduce a method that enables robots to learn continuously throughout their lifetime, rather than just once. Deep generative replay has emerged as a promising approach for continual learning in decision-making tasks. This approach addresses the problem of catastrophic forgetting by leveraging the generation of trajectories from previously encountered tasks to augment the current dataset. However, existing deep generative replay methods for continual learning rely on autoregressive models, which suffer from compounding errors in the generated trajectories. We propose a simple, scalable, and non-autoregressive method for continual learning in decision-making tasks using a generative model that generates task samples conditioned on the trajectory timestep. We evaluate our method on Continual World benchmarks and find that our approach achieves state-of-the-art performance on the average success rate metric among continual learning methods. The code used in these experiments is available at \href{https://github.com/WilliamYue37/t-DGR}{https://github.com/WilliamYue37/t-DGR}.

\newpage

\section{Introduction}
Continual learning, also known as lifelong learning, is a critical challenge in the advancement of general artificial intelligence, as it enables models to learn from a continuous stream of data encompassing various tasks, rather than having access to all data at once \citep{Ring:1994}. However, a major challenge in continual learning is the phenomenon of catastrophic forgetting, where previously learned skills are lost when attempting to learn new tasks \citep{MCCLOSKEY1989109}.

To mitigate catastrophic forgetting, replay methods have been proposed, which involve saving data from previous tasks and replaying it to the learner during the learning of future tasks. This approach mimics how humans actively prevent forgetting by reviewing material for tests and replaying memories in dreams. However, storing data from previous tasks requires significant storage space and becomes computationally infeasible as the number of tasks increases.

In the field of cognitive neuroscience, the Complementary Learning Systems theory offers insights into how the human brain manages memory. This theory suggests that the brain employs two complementary learning systems: a fast-learning episodic system and a slow-learning semantic system \citep{mcclelland1995complementary, KUMARAN2016512, cls}. The hippocampus serves as the episodic system, responsible for storing specific memories of unique events, while the neocortex functions as the semantic system, extracting general knowledge from episodic memories and organizing it into abstract representations \citep{4f488fef93a149c3894ef77d95fc1653}.

\begin{figure*}[t!]
\includegraphics[width=\linewidth]{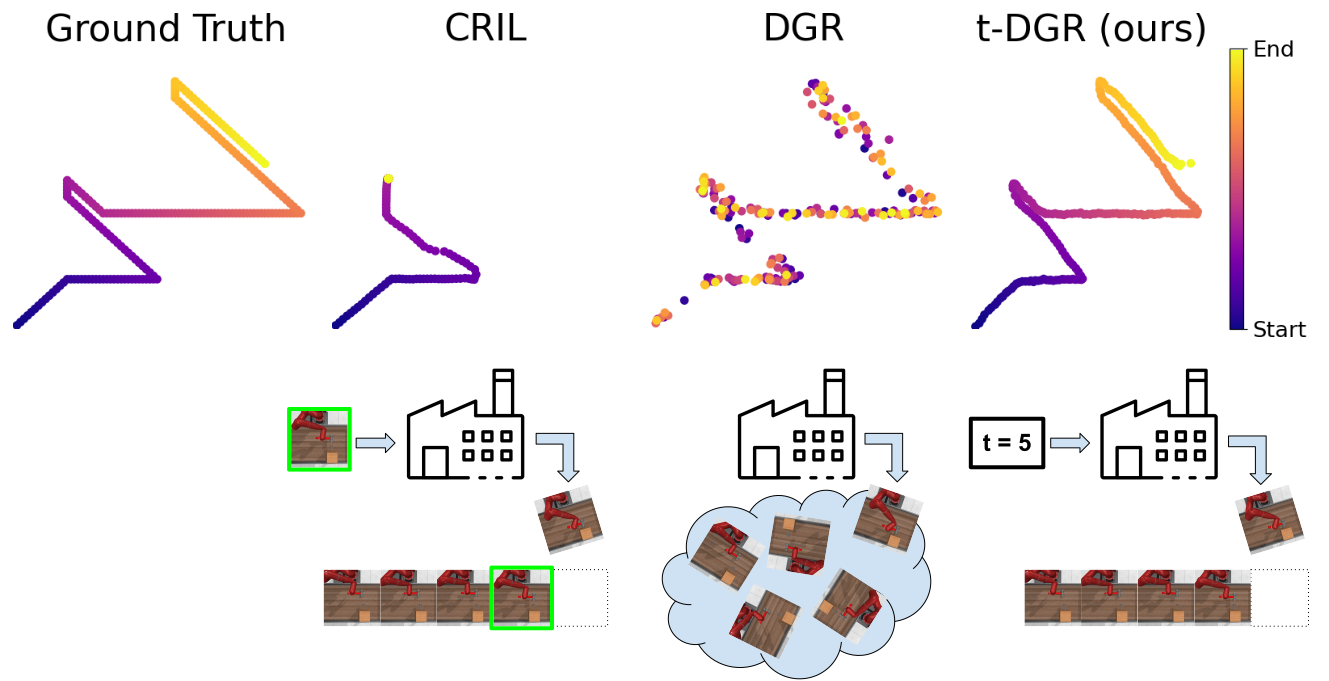}
\caption{The first row presents a comparison of three generative methods for imitating an agent's movement in a continuous 2D plane with Gaussian noise. The objective is to replicate the ground truth path, which transitions from darker to lighter colors. The autoregressive method (CRIL) encounters a challenge at the first sharp turn as nearby points move in opposing directions. Once the autoregressive method deviates off course, it never recovers and compromises the remaining trajectory. In contrast, sampling individual state observations i.i.d. without considering the temporal nature of trajectories (DGR) leads to a fragmented path with numerous gaps. Our proposed method t-DGR samples individual state observations conditioned on the trajectory timestep. By doing so, t-DGR successfully avoids the pitfalls of CRIL and DGR, ensuring a more accurate replication of the desired trajectory. The second row illustrates how each method generates trajectory data. CRIL generates the next state observation conditioned on the previous state observation. DGR, in contrast, does not attempt to generate a trajectory but generates individual state observations i.i.d. On the other hand, t-DGR generates state observations conditioned on the trajectory timestep.}
\label{fig:toy}
\end{figure*}

Drawing inspiration from the human brain, deep generative replay (DGR) addresses the catastrophic forgetting issue in decision-making tasks by using a generative model as the hippocampus to generate trajectories from past tasks and replay them to the learner which acts as the neocortex (Figure~\ref{fig:dgr}) \citep{shin2017continual}. The time-series nature of trajectories in decision-making tasks sets it apart from continual supervised learning, as each timestep of the trajectory requires sufficient replay. In supervised learning, the learner's performance is not significantly affected if it performs poorly on a small subset of the data. However, in decision-making tasks, poor performance on any part of the trajectory can severely impact the overall performance. Therefore, it is crucial to generate state-action pairs that accurately represent the distribution found in trajectories. Furthermore, the high-dimensional distribution space of trajectories makes it computationally infeasible to generate complete trajectories all at once.

Existing DGR methods adopt either the generation of individual state observations i.i.d. without considering the temporal nature of trajectories or autoregressive trajectory generation. Autoregressive approaches generate the next state(s) in a trajectory by modeling the conditional probability of the next state(s) given the previously generated state(s). However, autoregressive methods suffer from compounding errors in the generated trajectories. On the other hand, generating individual state observations i.i.d. leads to a higher sample complexity compared to generating entire trajectories, which becomes significant when replay time is limited (see Section~\ref{sec:rehearsal}).

To address the issues in current DGR methods, we propose a simple, scalable, and non-autoregressive trajectory-based DGR method. We define a generated trajectory as temporally coherent if the transitions from one state to the next appear realistic (refer to Section~\ref{sec:notation} for a formal definition). Given that imitation learning methods and many reinforcement learning methods are trained on single transition samples (e.g. state-action pairs), we do not require trajectories to exhibit temporal coherence. Instead, our focus is on ensuring an equal number of samples generated at each timestep of the trajectory to accurately represent the distribution found in trajectories. To achieve equal sample coverage at each timestep, we train our generator to produce state observations conditioned on the trajectory timestep, and then sample from the generator conditioned on each timestep of the trajectory. The intuition behind our method is illustrated in Figure~\ref{fig:toy}.

To evaluate the effectiveness of our proposed method, t-DGR, we conducted experiments on the Continual World benchmarks CW10 and CW20 \citep{wołczyk2021continual} using imitation learning. Our results indicate that t-DGR achieves state-of-the-art performance in terms of average success rate when compared to other top continual learning methods.

\section{Related Work}
This section provides an overview of existing continual learning methods, with a particular focus on pseudo-rehearsal methods.

\subsection{Continual Learning in the Real World}
As the field of continual learning continues to grow, there is an increasing emphasis on developing methods that can be effectively applied in real-world scenarios \citep{wang2024comprehensive, NEURIPS2019_e562cd9c, bang2021rainbow, hsu2019reevaluating, vandeven2019scenarios}. The concept of ``General Continual Learning" was introduced by \citet{buzzega2020dark} to address certain properties of the real world that are often overlooked or ignored by existing continual learning methods. Specifically, two important properties, bounded memory and blurry task boundaries, are emphasized in this work. Bounded memory refers to the requirement that the memory footprint of a continual learning method should remain bounded throughout the entire lifespan of the learning agent. This property is crucial to ensure practicality and efficiency in real-world scenarios. Additionally, blurry task boundaries highlight the challenge of training on tasks that are intertwined, without clear delineation of when one task ends and another begins. Many existing methods fail to account for this characteristic, which is common in real-world learning scenarios. While there are other significant properties associated with continual learning in the real world, this study focuses on the often-neglected aspects of bounded memory and blurry task boundaries. By addressing these properties, we aim to develop methods that are more robust and applicable in practical settings.

\subsection{Continual Learning Methods}
Continual learning methods for decision-making tasks can be categorized into three main categories.

\paragraph{Regularization}
Regularization methods in continual learning focus on incorporating constraints during model training to promote the retention of past knowledge. One simple approach is to include an $L_2$ penalty in the loss function. Elastic Weight Consolidation (EWC) builds upon this idea by assigning weights to parameters based on their importance for previous tasks using the Fisher information matrix \citep{Kirkpatrick_2017}. MAS measures the sensitivity of parameter changes on the model's output, prioritizing the retention of parameters with a larger effect \citep{aljundi2018memory}. VCL leverages variational inference to minimize the Kullback-Leibler divergence between the current and prior parameter distributions \citep{nguyen2018variational}. Progress and Compress learns new tasks using a separate model and subsequently distills this knowledge into the main model while safeguarding the previously acquired knowledge \citep{schwarz2018progress}. However, regularization methods may struggle with blurry task boundaries as they rely on knowledge of task endpoints to apply regularization techniques effectively. In our experiments, EWC was chosen as the representative regularization method based on its performance in the original Continual World experiments \citep{wołczyk2021continual}.

\paragraph{Architecture-based Methods}
Architecture-based methods aim to maintain distinct sets of parameters for each task, ensuring that future learning does not interfere with the knowledge acquired from previous tasks. Packnet \citep{mallya2018packnet}, UCL \citep{NEURIPS2019_2c3ddf4b}, and AGS-CL \citep{jung2021continual} all safeguard previous task information in a neural network by identifying important parameters and freeing up less important parameters for future learning. Identification of important parameters can be done through iterative pruning (Packnet), parameter uncertainty (UCL), and activation value (AGS-CL). However, a drawback of parameter isolation methods is that each task requires its own set of parameters, which may eventually exhaust the available parameters for new tasks and necessitate a dynamically expanding network without bounded memory \citep{yoon2018lifelong}. Additionally, parameter isolation methods require training on a single task at a time to prune and isolate parameters, preventing concurrent learning from multiple interwoven tasks. In our experiments, PackNet was selected as the representative architecture-based method based on its performance in the original Continual World experiments \citep{wołczyk2021continual}.

\begin{figure*}
% \includesvg[width=1\linewidth]{figs/t-DGR.svg}
\includegraphics[width=\linewidth]{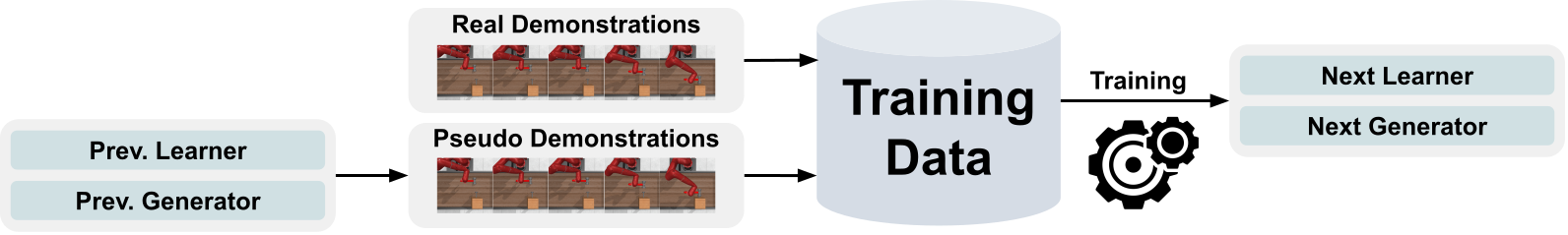}
\caption{The deep generative replay paradigm. The algorithm learns to generate trajectories from past tasks to augment real trajectories from the current task in order to mitigate catastrophic forgetting. Both the generator and policy model are updated with this augmented dataset.}
\label{fig:dgr}
\end{figure*}

\paragraph{Pseudo-rehearsal Methods}
\label{sec:rehearsal}
Pseudo-rehearsal methods mitigate the forgetting of previous tasks by generating synthetic samples from past tasks and replaying them to the learner. Deep generative replay (DGR) (Figure~\ref{fig:dgr}) utilizes a generative model, such as generative adversarial networks \citep{goodfellow2014generative}, variational autoencoders \citep{kingma2022autoencoding}, or diffusion models \citep{ho2020denoising, Sohl}, to generate the synthetic samples. Originally, deep generative replay was proposed to address continual supervised learning problems, where the generator only needed to generate single data point samples \citep{shin2017continual}. However, in decision-making tasks, expert demonstrations consist of trajectories (time-series) with a significantly higher-dimensional distribution space.

One existing DGR method generates individual state observations i.i.d. instead of entire trajectories. However, this approach leads to a higher sample complexity compared to generating entire trajectories. The sample complexity of generating enough individual state observations i.i.d. to cover every portion of the trajectory $m$ times can be described using the Double Dixie Cup problem \citep{doubleDixieCup}. For trajectories of length $n$, it takes an average of $\Theta(n \log n + m n \log\log n)$ i.i.d. samples to ensure at least $m$ samples for each timestep. In scenarios with limited replay time (small $m$) and long trajectories (large $n$) the sample complexity can be approximated as $\Theta(n \log n)$ using the Coupon Collector's problem \citep{couponCollector}.  The additional $\Theta(\log n)$ factor reduces the likelihood of achieving complete sample coverage of the trajectory when the number of replays or replay time is limited, especially considering the computationally expensive nature of current generative methods. Furthermore, there is a risk that the generator assigns different probabilities to each timestep of the trajectory, leading to a selective focus on certain timesteps rather than equal representation across the trajectory.

Another existing DGR method is autoregressive trajectory generation. In the existing autoregressive method, CRIL, a generator is used to generate samples of the initial state, and a dynamics model predicts the next state based on the current state and action \citep{gao2021cril}.  However, even with a dynamics model accuracy of 99\% and a 1\% probability of deviating from the desired trajectory, the probability of an autoregressively generated trajectory going off course is $1 - 0.99^n$, where $n$ denotes the trajectory length. With a trajectory length of $n = 200$ (as used in our experiments), the probability of an autoregressively generated trajectory going off course is $1 - 0.99^{200} = 0.87$. This example demonstrates how the issue of compounding error leads to a high probability of failure, even with a highly accurate dynamics model.

In our experiments, t-DGR is evaluated against all existing trajectory generation methods in pseudo-rehearsal approaches to assess how well t-DGR addresses the limitations of those methods.

\section{Background}
This section introduces notation and the formulation of the continual imitation learning problem that we use in this chapter. Additionally, we provide a concise overview of diffusion probabilistic models used in our generative model implementation.

\subsection{Imitation Learning}
Imitation learning algorithms aim to learn a policy $\pi_\theta$ parameterized by $\theta$ by imitating a set of expert demonstrations $D = \{\tau_i\}_{i = 1 \ldots M}$. Each trajectory $\tau_i$ consists of a sequence of state-action pairs $\{(s_j, a_j)\}_{j = 1 \ldots |\tau_i|}$ where $|\tau_i|$ is the length of the trajectory. Each trajectory comes from a task $\mathcal{T}$ which is a Markov decision process that can be represented as a tuple $\langle S, A, T, \rho_0 \rangle$ with state space $S$, action space $A$, transition dynamics $T: S \times A \times S \to [0, 1]$, and initial state distribution $\rho_0$. Various algorithms exist for imitation learning, including behavioral cloning, GAIL \citep{ho2016generative}, and inverse reinforcement learning \citep{ng2000inverse}. In this work, we use behavioral cloning where the objective can be formulated as minimizing the loss function: 
\begin{equation} \label{eq:mseloss}
\mathcal{L}(\theta) = \mathbb{E}_{s,a \sim D}\bigg[\big\|\pi_\theta(s) - a\big\|^2_2 \bigg]
\end{equation}
where the state and action spaces are continuous.

\subsection{Continual Imitation Learning}
\label{sec:CL}
In the basic formulation most common in the field today, continual imitation learning involves sequentially solving multiple tasks $\mathcal{T}_1, \mathcal{T}_2, \ldots, \mathcal{T}_N$. When solving for task $\mathcal{T}_i$, the learner only gets data from task $\mathcal{T}_i$ and can not access data for any other task. In a more general scenario, certain tasks may have overlapping boundaries, allowing the learner to encounter training data from multiple tasks during certain phases of training. The learner receives a continuous stream of training data in the form of trajectories $\tau_1, \tau_2, \tau_3, \ldots$ from the environment, where each trajectory $\tau$ corresponds to one of the $N$ tasks. However, the learner can only access a limited contiguous portion of this stream at any given time. 

Let $s_i$ be the success rate of task $\mathcal{T}_i$ after training on all $N$ tasks. The continual imitation learning objective is defined as maximizing the average success rate over all tasks: 
\begin{equation} \label{eq:metric}
S = \frac{1}{N}\sum_{i = 1}^N s_i
\end{equation}
The primary issue that arises from the continual learning problem formulation is the problem of catastrophic forgetting where previously learned skills are forgotten when training on a new task. 

\subsection{Diffusion Probabilistic Models}
Diffusion probabilistic models \citep{ho2020denoising, Sohl} generate data through a learned reverse denoising diffusion process $p_\theta(x_{t - 1} \mid x_t)$. The forward diffusion process $q(x_t \mid x_{t - 1})$ gradually adds Gaussian noise to an input $x_0$ at each time step $t$, ultimately resulting in pure noise $x_T$ at $t = T$. The forward diffusion process is defined as:
\begin{equation}
    q(x_t \mid x_{t - 1}) = \mathcal{N}\left(x_t; \sqrt{1 - \beta_t}x_{t - 1}, \beta_t\mathbf{I}\right)
\end{equation}
where $0 < \beta_t < 1$ is defined by a known variance schedule. In our implementation, we adopted the cosine schedule proposed by \citet{nichol2021improved}. For a sufficiently large time horizon $T$ and a well-behaved variance schedule, $x_T$ approximates an isotropic Gaussian distribution. If we had the reverse diffusion process $p(x_{t - 1} \mid x_t)$, we could sample $x_T \sim \mathcal{N}(0, \mathbf{I})$ and obtain a sample from $q(x_0)$ by denoising $x_T$ with $p(x_{t - 1} \mid x_t)$. However, computing $p(x_{t - 1} \mid x_t)$ is intractable as it necessitates knowledge of the distribution of all possible $x_t$. Instead, we approximate $p(x_{t - 1} \mid x_t)$ using a neural network:
\begin{equation} \label{eq:ptheta}
    p_\theta(x_{t - 1} \mid x_t) = \mathcal{N}\left(x_{t - 1}; \mu_\theta(x_t, t), \Sigma(x_t, t)\right) 
\end{equation}
Since $q$ and $p_\theta$ can be viewed as a variational auto-encoder \citep{kingma2022autoencoding}, we can use the variational lower bound to minimize the negative log-likelihood of the reverse process. We can express $\mu_\theta(x_t, t)$ from Equation~\ref{eq:ptheta} as:
\begin{equation}
    \mu_\theta(x_t, t) = \frac{1}{\sqrt{\alpha_t}}\left(x_t - \frac{\beta_t}{\sqrt{1 - \overline{\alpha}_t}}\epsilon_\theta(x_t, t)\right)
\end{equation}
where $\alpha_t = 1 - \beta_t$ and $\overline{\alpha}_t = \prod_{s = 0}^t \alpha_s$. The training loss can then be defined as:
\begin{equation} \label{eq:diffLoss}
    \mathcal{L}(\theta) = \mathbb{E}_{x_0, t, \epsilon}\left[\lVert\epsilon - \epsilon_\theta(x_t, t)\rVert^2\right]
\end{equation}
Note that the timesteps $t$ in the diffusion process differ from the trajectory timesteps $t$. Henceforth, we will refer only to the trajectory timesteps $t$.

\subsection{Notation}
\label{sec:notation}
Deep generative replay involves training two models: a generator $G_\gamma$ parameterized by $\gamma$ and a learner $\pi_\theta$ parameterized by $\theta$. We define $G_\gamma^{(i)}$ as the generator trained on tasks $\mathcal{T}_1 \ldots \mathcal{T}_i$ and capable of generating data samples from tasks $\mathcal{T}_1 \ldots \mathcal{T}_i$. Similarly, $\pi_\theta^{(i)}$ represents the learner trained on tasks $\mathcal{T}_1 \ldots \mathcal{T}_i$ and able to solve tasks $\mathcal{T}_1 \ldots \mathcal{T}_i$.

A sequence of state observations $(s_1, s_2, \ldots, s_{n - 1}, s_n)$ is \textbf{temporally coherent} if $\forall 1 \leq i < n, \exists a \in A : T(s_i, a, s_{i + 1}) > \varepsilon$, where $0 < \varepsilon < 1$ is a small constant representing a threshold for negligible probabilities.

\section{Method}
Our proposed method, t-DGR, tackles the challenge of generating long trajectories by training a generator, denoted as $G_\gamma(j)$, which is conditioned on the trajectory timestep $j$ to generate state observations. Pseudocode for t-DGR is provided as Algorithm~\ref{alg:t-DGR}. The algorithm begins by initializing the task index, replay ratio, generator model, learner model, and learning rates (Line \ref{line:init}). The replay ratio, denoted as $0 \leq r < 1$, determines the percentage of training samples seen by the learner that are generated. Upon receiving training data from the environment, t-DGR calculates the number of trajectories to generate based on the replay ratio $r$ (Lines \ref{line:data}-\ref{line:n}). The variable $L$ (Line \ref{line:L}) represents the maximum length of trajectories observed so far.

To generate a trajectory $\tau$ of length $L$, t-DGR iterates over each timestep $1 \leq j \leq L$ (Line \ref{line:forL}). At each timestep, t-DGR generates the $j$-th state observation of the trajectory using the previous generator $G_\gamma^{(t - 1)}$ conditioned on timestep $j$ (Line \ref{line:gen}), and then labels it with an action using the previous policy $\pi_\theta^{(t - 1)}$ (Line \ref{line:label}). After generating all timesteps in the trajectory $\tau$, t-DGR adds it to the existing training dataset (Line \ref{line:add}). Note that the generated state observations within a trajectory do not have temporal coherence, as each state observation is generated independently of other timesteps. This approach is acceptable since our learner is trained on state-action pairs rather than full trajectories. However, unlike generating state observations i.i.d., our method ensures equal coverage of every timestep during the generative process, significantly reducing sample complexity.

Once t-DGR has augmented the training samples from the environment with our generated training samples, t-DGR employs backpropagation to update both the generator and learner using the augmented dataset (Lines \ref{line:beginBP}-\ref{line:endBP}). The t-DGR algorithm continues this process of generative replay throughout the agent's lifetime, which can be infinite (Line \ref{line:loop}). Although we perform the generative process of t-DGR at task boundaries for ease of understanding, no part of t-DGR is dependent on clear task boundaries.

\paragraph{Architecture}
We employ a U-net \citep{ronneberger2015u} trained on the loss specified in Equation~\ref{eq:diffLoss} to implement the generative diffusion model $G_\gamma$. Since we utilize proprioceptive observations in our experiments, $\pi_\theta$ is implemented with a multi-layer perceptron trained on the loss specified in Equation~\ref{eq:mseloss}.

\begin{algorithm}[t!]
\caption{Trajectory-based Deep Generative Replay (t-DGR)}
\label{alg:t-DGR}
\begin{algorithmic}[1]
\State Initialize task index $t = 0$, replay ratio $r$, generator $G^{(0)}_\gamma$, learner $\pi^{(0)}_\theta$, and learning rates $\lambda_\gamma, \lambda_\theta$. \label{line:init}
\While{new task available} \label{line:loop}
    \State $t \gets t + 1$
    \State Initialize dataset $D$ with trajectories from task $t$. \label{line:data}
    \State $n \gets \frac{r * |D|}{1 - r}$ \Comment{number of trajectories to generate} \label{line:n}
    \For{$i = 1$ to $n$}
        \State $L \gets$ maximum trajectory length \label{line:L}
        \State $\tau \gets \emptyset$ \Comment{initialize trajectory of length $L$}
        \For{$j = 1$ to $L$} \label{line:forL}
        \State $S \gets G_\gamma^{(t - 1)}(j)$ \Comment{generate states} \label{line:gen}
        \State $A \gets \pi_\theta^{(t - 1)}(S)$ \Comment{label with actions} \label{line:label}
        \State $\tau_j \gets (S, A)$ \Comment{add to trajectory}
        \EndFor
        \State $D \gets D \cup \tau$ \Comment{add generated trajectory to $D$} \label{line:add}
    \EndFor
    \State Update generator and learner using $D$ \label{line:beginBP}
    \State $\gamma^{(t)} \gets \gamma^{(t - 1)} - \lambda_\gamma \nabla_\gamma \mathcal{L}_{G^{(t - 1)}}(\gamma^{(t - 1)})$
    \State $\theta^{(t)} \gets \theta^{(t - 1)} - \lambda_\theta \nabla_\theta \mathcal{L}_{\pi^{(t - 1)}}(\theta^{(t - 1)})$ \label{line:endBP}
\EndWhile
\end{algorithmic}
\end{algorithm}

\section{Experiments}
In this section, we outline the experimental setup and performance metrics employed to compare t-DGR with representative methods, followed by an analysis of experimental results across different benchmarks and performance metrics.

\subsection{Experimental Setup}
We evaluate our method on the Continual World benchmarks CW10 and CW20 \citep{wołczyk2021continual}, along with our own variant of CW10 called BB10 that evaluates the ability of methods to handle blurry task boundaries. CW10 consists of a sequence of 10 Meta-World \citep{yu2021metaworld} tasks, where each task involves a Sawyer arm manipulating one or two objects in the Mujoco physics simulator. For computational efficiency, we provide the agents with proprioceptive observations. Notably, the observation and action spaces are continuous and remain consistent across all tasks. CW20 is an extension of CW10 with the tasks repeated twice. To our knowledge, Continual World is the only standard continual learning benchmark for decision-making tasks. BB10 gives data to the learner in 10 sequential buckets $B_1, \ldots, B_{10}$. Data from task $\mathcal{T}_i$ from CW10 is split evenly between buckets $B_{i - 1}$, $B_i$, and $B_{i + 1}$, except for the first and last task. Task $\mathcal{T}_1$ is evenly split between buckets $B_0$ and $B_1$, and task $\mathcal{T}_{10}$ is evenly split between buckets $B_9$ and $B_{10}$.

In order to bound memory usage, task conditioning should utilize fixed-size embeddings like natural language task embeddings, rather than maintaining a separate final neural network layer for each individual task. For simplicity, we condition the model on the task with a one-hot vector as a proxy for natural language prompt embeddings. For BB10, the model is still conditioned on the underlying task rather than the bucket. Additionally, we do not allow separate biases for each task, as originally done in EWC \citep{Kirkpatrick_2017}. Expert demonstrations for training are acquired by gathering 100 trajectories per task using hand-designed policies from Meta-World, with each trajectory limited to a maximum of 200 steps. Importantly, the learner model remains consistent across different methods and benchmark evaluations. Moreover, we maintain a consistent replay ratio of $r = 0.9$ across all pseudo-rehearsal methods. 

We estimated the success rate $S$ of a model by running each task 100 times. The metrics for each method were computed using 5 seeds to create a 90\% confidence interval. Further experimental details, such as hyperparameters, model architecture, random seeds, and computational resources, are included in Appendix~\ref{app:experiment},~\ref{app:hyperparams},~\ref{app:model_arch}. This standardization enables a fair and comprehensive comparison of our proposed approach with other existing methods.

\subsection{Metrics}
We evaluate our models using three metrics proposed by the Continual World benchmark \citep{wołczyk2021continual}, with the average success rate being the primary metric. Although the forward transfer and forgetting metrics are not well-defined in a setting with blurry task boundaries, they are informative within the context of Continual World benchmarks. As a reminder from Section~\ref{sec:CL}, let $N$ denote the number of tasks, and $s_i$ represent the success rate of the learner on task $\mathcal{T}_i$. Additionally, let $s_i(t)$ denote the success rate of the learner on task $\mathcal{T}_i$ after training on tasks $\mathcal{T}_1$ to $\mathcal{T}_t$.

\paragraph{Average Success Rate} The average success rate, as given by Equation~\ref{eq:metric}, serves as the primary evaluation metric for continual learning methods.

\paragraph{Average Forward Transfer} We introduce a slightly modified metric for forward transfer that applies to a broader range of continual learning problems beyond just continual reinforcement learning in the Continual World benchmark. Let $s_i^{\mathrm{ref}}$ represent the reference performance of a single-task experiment on task $\mathcal{T}_i$. The forward transfer metric $FT_i$ is computed as follows:
\begin{align*}
FT_i = \frac{D_i - D_i^{\mathrm{ref}}}{1 - D_i^{\mathrm{ref}}} && D_i = \frac{s_i(i) + s_i(i - 1)}{2} && D_i^{\mathrm{ref}} = \frac{s_i^{\mathrm{ref}}}{2}
\end{align*}
The expressions for $D_i$ and $D^{\mathrm{ref}}_i$ serve as approximations of the integral of task $\mathcal{T}_i$ performance with respect to the training duration for task $\mathcal{T}_i$. The average forward transfer $FT$ is then defined as the mean forward transfer over all tasks, calculated as $FT = \frac{1}{N}\sum_{i = 1}^N FT_i$.

\paragraph{Average Forgetting} We measure forgetting using the metric $F_i$, which represents the amount of forgetting for task $i$ after all training has concluded. $F_i$ is defined as the difference between the success rate on task $\mathcal{T}_i$ immediately after training and the success rate on task $\mathcal{T}_i$ at the end of training. \[F_i = s_i(i) - s_i(N)\] The average forgetting $F$ is then computed as the mean forgetting over all tasks, given by $F = \frac{1}{N}\sum_{i = 1}^N F_i$.

\subsection{Baselines}
We compare the following methods on the Continual World benchmark using average success rate as the primary evaluation metric. Representative methods were chosen based on their success in the original Continual World experiments, while DGR-based methods were selected to evaluate whether t-DGR addresses the limitations of existing pseudo-rehearsal methods.
\begin{itemize}
    \item \textbf{Finetune:} The policy is trained only on data from the current task.
    \item \textbf{Multitask:} The policy is trained on data from all tasks simultaneously.
    \item \textbf{oEWC~\citep{schwarz2018progress}:} A variation of EWC known as online Elastic Weight Consolidation (oEWC) bounds the memory of EWC by employing a single penalty term for the previous model instead of individual penalty terms for each task. This baseline is the representative regularization-based method.
    \item \textbf{PackNet~\citep{mallya2018packnet}:} This baseline is the representative parameter isolation method. Packnet safeguards previous task information in a neural network by iteratively pruning, freezing, and retraining parts of the network.
    \item \textbf{DGR~\citep{shin2017continual}:} This baseline is a deep generative replay method that only generates individual state observations i.i.d. and not entire trajectories.
    \item \textbf{CRIL~\citep{gao2021cril}:} This baseline is a deep generative replay method that trains a policy along with a start state generator and a dynamics model that predicts the next state given the current state and action. Trajectories are generated by using the dynamics model and policy to autoregressively generate next states from a start state.
    \item \textbf{t-DGR:} Our proposed method.
\end{itemize}
Due to the inability of oEWC and PackNet to handle blurry task boundaries, we made several adjustments for CW20 and BB10. Since PackNet cannot continue training parameters for a task once they have been fixed, we treated the second repetition of tasks in CW20 as distinct from the first iteration, resulting in PackNet being evaluated with $N = 20$, while the other methods were evaluated with $N = 10$. As for BB10 and its blurry task boundaries, the best approach we could adopt with oEWC and PackNet was to apply their regularization techniques at regular training intervals rather than strictly at task boundaries. During evaluation, all tasks were assessed using the last fixed set of parameters in the case of PackNet.

\subsection{Ablations}

To evaluate the effect of generator architecture on pseudo-rehearsal methods, we included an ablation where the diffusion models in pseudo-rehearsal methods are replaced with Wasserstein generative adversarial networks (GAN) with gradient penalty \citep{gulrajani2017improved}. To evaluate the effectiveness of diffusion models at generative replay, we evaluated the quality of generated samples for past tasks at all stages of continual learning. Since we utilize proprioceptive observations rather than RGB camera observations, there is no practical way to qualitatively evaluate the generated samples. The proprioceptive observations are 39-dimensional, comprising 3D positions and quaternions of objects, and the degree to which the robot gripper is open \citep{yu2021metaworld}. These raw numbers cannot be qualitatively evaluated like RGB images. Instead, we use the average diffusion loss as a quantitative proxy metric for generation quality. When the generator model is learning task $i$, we compute the average diffusion loss for data samples from tasks 1 to $i - 1$. We then compare this average diffusion loss to the average diffusion loss when the generator model first learned task $i$. The difference between the two provides an estimate of how much generative replay has degraded the generation quality of previous tasks.

\newcommand{\fs}[1]{\footnotesize $\pm$#1}

\begin{table}[h]
    \captionsetup[subtable]{font=large}
  \centering
  
  \resizebox{0.7\textwidth}{!}{
  \begin{subtable}{\textwidth}
      \centering
      \caption{CW10}
      \begin{tabular}{l c c c c}
        \toprule
        Method & Success Rate $\uparrow$ & FT$\uparrow$ & Forgetting$\downarrow$ \\
        \midrule
        Finetune & 16.4 \fs{6.4} & -3.0 \fs{6.0} & 78.8 \fs{7.6} \\
        Multitask & 97.0 \fs{1.0} & N/A & N/A \\
        \midrule
        oEWC & 18.6 \fs{5.3} & -6.3 \fs{5.7} & 74.1 \fs{6.1} \\
        PackNet & 81.4 \fs{3.7} & -14.8 \fs{7.8} & \textbf{-0.1} \fs{1.2} \\
        DGR (gan) & 21.2 \fs{3.7} & -2.6 \fs{3.3} & 74.4 \fs{4.7} \\
        CRIL (gan) & 24.0 \fs{4.4} & 0.3 \fs{4.5} & 72.0 \fs{5.4} \\
        t-DGR (gan) & 17.4 \fs{4.4} & 0.1 \fs{3.1} & 79.8 \fs{4.2} \\
        DGR (diffusion) & 75.0 \fs{5.8} & -4.3 \fs{5.1} & 17.8 \fs{4.1} \\
        CRIL (diffusion) & 28.4 \fs{10.6} & -1.1 \fs{2.8} & 68.6 \fs{10.4} \\
        t-DGR (diffusion) & \textbf{81.9} \fs{3.3} & \textbf{-0.3} \fs{4.9} & 14.4 \fs{2.5} \\
        \bottomrule
      \end{tabular}
      
    \label{tbl:CW10}
  \end{subtable}
  }

  \vspace{0.5cm} % Adjust vertical spacing between the tables

  \resizebox{0.7\textwidth}{!}{
  \begin{subtable}{\textwidth}
      \centering
      \caption{BB10}
      \begin{tabular}{l c}
        \toprule
        Method & Success Rate $\uparrow$ \\
        \midrule
        Finetune & 21.7 \fs{2.6} \\
        Multitask & 97.0 \fs{1.0} \\
        \midrule
        oEWC & 21.8 \fs{1.7} \\
        PackNet & 26.9 \fs{5.6} \\
        DGR (gan) & 35.6 \fs{4.3} \\
        CRIL (gan) & 37.5 \fs{4.7} \\
        t-DGR (gan) & 28.3 \fs{3.9} \\
        DGR (diffusion) & 75.3 \fs{4.4} \\
        CRIL (diffusion) & 53.5 \fs{5.5} \\
        t-DGR (diffusion) & \textbf{81.7}    \fs{4.0} \\
        \bottomrule
      \end{tabular}
      
    \label{tbl:BB10}
  \end{subtable}
  }

  \vspace{0.5cm} % Adjust vertical spacing between the tables

    \resizebox{0.7\textwidth}{!}{
  \begin{subtable}{\textwidth}
      \centering
      \caption{CW20}
      \begin{tabular}{l c c c c}
        \toprule
        Method & Success Rate $\uparrow$ & FT$\uparrow$ & Forgetting$\downarrow$ \\
        \midrule
        Finetune & 14.2 \fs{4.0} & -0.5 \fs{3.0} & 82.2 \fs{5.6} \\
        Multitask & 97.0 \fs{1.0} & N/A & N/A \\
        \midrule
        oEWC & 19.4 \fs{5.3} & -2.8 \fs{4.1} & 75.2 \fs{7.5} \\
        PackNet & 74.1 \fs{4.1} & -20.4 \fs{3.4} & \textbf{-0.2} \fs{0.9} \\
        DGR (gan) & 19.0 \fs{2.1} & 0.8 \fs{2.3} & 78.6 \fs{2.3} \\
        CRIL (gan) & 26.7 \fs{6.5} & -0.6 \fs{0.7} & 72.1 \fs{6.2} \\
        t-DGR (gan) & 20.6 \fs{6.1} & 0.7 \fs{3.4} & 76.3 \fs{5.3} \\
        DGR (diffusion) & 74.1 \fs{4.1} & 18.9 \fs{2.9} & 23.3 \fs{3.3} \\
        CRIL (diffusion) & 50.8 \fs{4.4} & 4.4 \fs{4.9} & 46.1 \fs{5.4} \\
        t-DGR (diffusion) & \textbf{83.9} \fs{3.0} & \textbf{30.6} \fs{4.5} & 14.6 \fs{2.9} \\
        \bottomrule
      \end{tabular}
      
        \label{tbl:CW20}
  \end{subtable}
  }

    \vspace{0.5cm} % Adjust vertical spacing between the tables

  \resizebox{0.7\textwidth}{!}{
  \begin{subtable}{\textwidth}
      \centering
      \caption{Replay Ratio}
      \begin{tabular}{c|c c c}
        \toprule
        Ratio & t-DGR & DGR \\
        \midrule
        0.5 & \textbf{63.2} \fs{2.6} & 52.8 \fs{2.9} \\
        0.6 & \textbf{66.3} \fs{4.4} & 56.9 \fs{4.5} \\
        0.7 & \textbf{70.8} \fs{4.1} & 62.5 \fs{3.6} \\
        0.8 & \textbf{75.0} \fs{6.9} & 69.2 \fs{4.9} \\
        0.9 & \textbf{81.9} \fs{3.3} & 75.0 \fs{5.8} \\
        \bottomrule
      \end{tabular}
      
        \label{tbl:ratio}
  \end{subtable}
  }
  
  \caption{Tables (a), (b), and (c) present the results for Continual World 10, Blurry Boundaries 10, and Continual World 20, respectively. The tables display the average success rate, forward transfer, and forgetting (if applicable) with 90\% confidence intervals using 5 random seeds. An up arrow indicates that higher values are better and a down arrow indicates that smaller values are better. Table (d) compares the impact of replay amount on the average success rate of t-DGR and DGR on CW10 with 90\% confidence intervals obtained using 5 random seeds. The best results are highlighted in bold.}
  \label{tab:all_tables}
\end{table}

\subsection{Discussion}
t-DGR emerges as the leading method, demonstrating the highest success rate on CW10 (Table~\ref{tbl:CW10}), CW20 (Table~\ref{tbl:CW20}), and BB10 (Table~\ref{tbl:BB10}). Notably, PackNet's performance on the second iteration of tasks in CW20 diminishes, highlighting its limited capacity for continually accommodating new tasks. This limitation underscores the fact that PackNet falls short of being a true lifelong learner, as it necessitates prior knowledge of the task count for appropriate parameter capacity allocation. On the contrary, pseudo-rehearsal methods, such as t-DGR, exhibit improved performance with the second iteration of tasks in CW20 due to an increased replay time. These findings emphasize the ability of DGR methods to effectively leverage past knowledge, as evidenced by their superior forward transfer in both CW10 and CW20.

BB10 (Table~\ref{tbl:BB10}) demonstrates that pseudo-rehearsal methods are mostly unaffected by blurry task boundaries, whereas PackNet's success rate experiences a significant drop-off. This discrepancy arises from the fact that PackNet's regularization technique does not work effectively with less clearly defined task boundaries.

In our experiments across CW10, CW20, and BB10, we observed that diffusion models outperform GANs as the generator for pseudo-rehearsal methods. We hypothesize that the distributional shifts in continual learning exacerbate instability issues with GAN training \citep{salimans2016improved, brock2018large, miyato2018spectral}. The motivation behind CRIL as stated in the paper \citep{gao2021cril} is to alleviate the burden of trajectory generation from the generator by transferring some of the generation complexity to a dynamics model. Our findings support this motivation when working with less capable generators \citep{dhariwal2021diffusion}, as reducing the generator's burden consistently enhances performance of GAN-based pseudo-rehearsal methods across all benchmarks. Among pseudo-rehearsal techniques, CRIL places the least demand on its generator, which only needs to produce the initial frame of a trajectory. In contrast, DGR requires generation of every frame, and t-DGR, the most demanding, requires generation of every frame along with the handling of trajectory timestep conditioning. Results from our GAN-based pseudo-rehearsal experiments on CW10, CW20, and BB10 indicate that CRIL outperforms DGR, which in turn outperforms t-DGR. However, experiments using a more capable diffusion generator \citep{dhariwal2021diffusion} reveal a reversal in performance ranking among these methods, suggesting that more capable generators diminish the need to offload generation complexity from the main generator.

The diminishing performance gap between DGR and t-DGR as the replay ratio increases in Table~\ref{tbl:ratio} indicates that a higher replay ratio reduces the likelihood of any portion of the trajectory being insufficiently covered when sampling individual state observations i.i.d., thereby contributing to improved performance. This trend supports the theoretical sample complexity of DGR derived in Section~\ref{sec:rehearsal}, as $\Theta(n \log n + m n \log\log n)$ closely approximates the sample complexity of t-DGR, $\Theta(mn)$, when the replay amount $m \to \infty$. However, while DGR can achieve comparable performance to t-DGR with a high replay ratio, the availability of extensive replay time is often limited in many real-world applications.

\begin{figure*}[h]
\includegraphics[width=\linewidth]{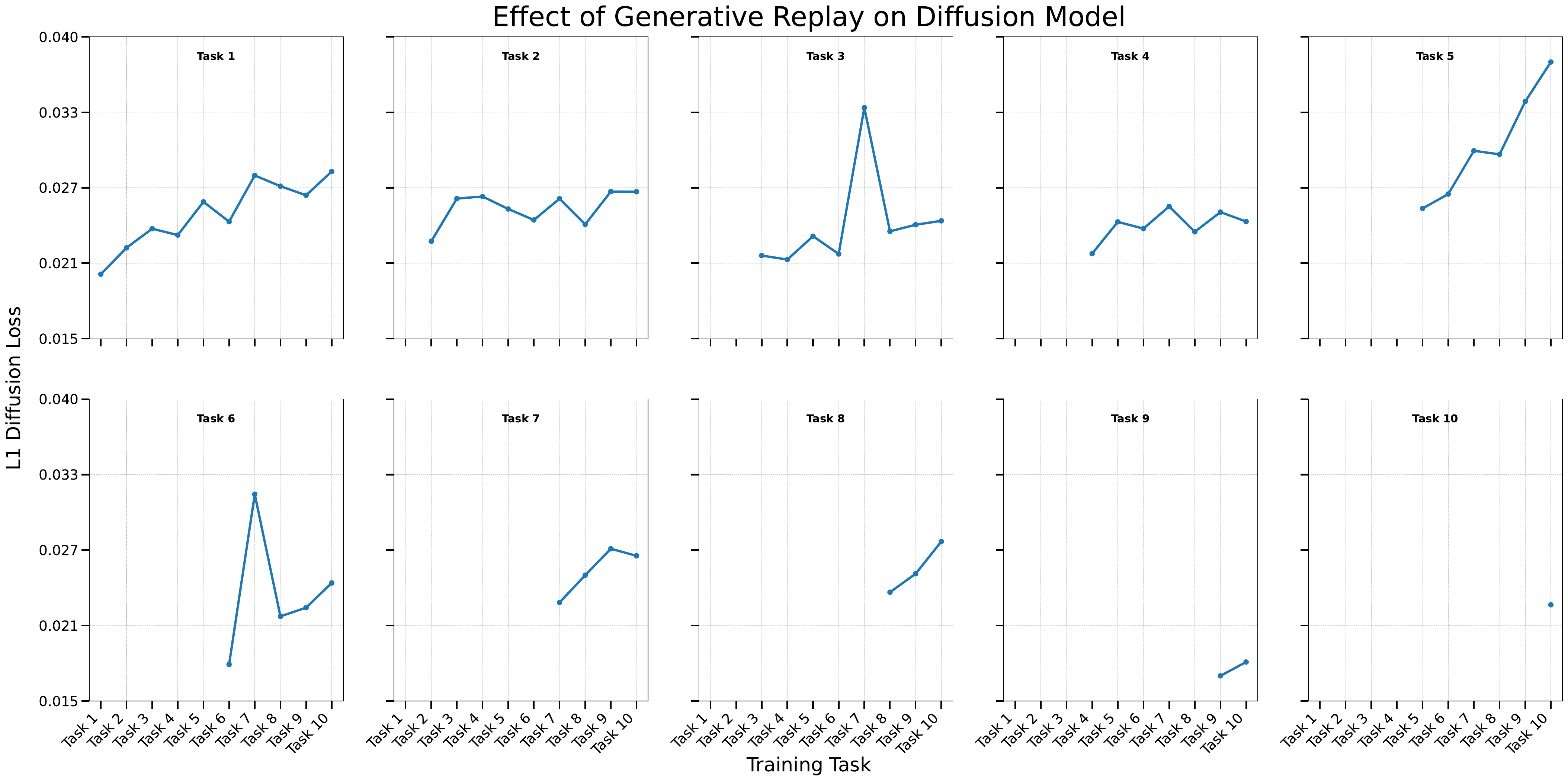}
\caption{This table illustrates the ability of the diffusion model in t-DGR to generate past data as it continues to learn additional tasks in CW10 through generative replay. The line plot for task $i$ plots the average diffusion loss of the diffusion model in future tasks on task $i$ data. The loss is an L1 version of the diffusion training loss in Equation~\ref{eq:diffLoss}.}
\label{fig:curve}
\end{figure*}

Recent studies have indicated that applying generative replay to diffusion models for image data leads to a collapse of generation quality due to compounding errors in the generated synthetic data \citep{Zajac2023ExploringCL, masip2023continual, smith2023continual}. However, our experiments reveal that when generating lower dimensional proprioceptive data, the diffusion model is able to maintain generation quality. Although Figure~\ref{fig:curve} shows that generative replay degrades generation quality of past tasks as the diffusion model learns new tasks, the success rates in Table~\ref{tab:all_tables} suggest that this degradation in generation quality is not severe enough to impact the learner's performance. Notably, generation quality for past tasks appears to deteriorate in tasks 5 and 7 but improves in the subsequent task, suggesting that the error compounding in diffusion models during generative replay might not be as severe as previously assumed, particularly for lower-dimensional data.

Overall, t-DGR exhibits promising results, outperforming other methods in terms of success rate in all evaluations. Notably, t-DGR achieves a significant improvement over existing pseudo-rehearsal methods on CW20 using a Welch t-test with a significance level of $\text{p-value} = 0.005$. Its ability to handle blurry task boundaries, leverage past knowledge, and make the most of replay opportunities position it as a state-of-the-art method for continual lifelong learning in decision-making.

\chapter{Learning to Use Memory}
\label{chp:memory}

In Chapter~\ref{chp:lifelong}, we present a method to enhance lifelong learning for robots operating over extended time horizons. However, continuous learning over a lifetime requires reflecting on past experiences. Humans often learn by drawing on past memories, whether by recalling a poor decision to avoid repeating it or revisiting a long-forgotten math concept from high school. This chapter introduces a framework enabling humans to teach agents how to effectively utilize their memory mechanisms. In Partially Observable Markov Decision Processes, integrating an agent's history into memory poses a significant challenge for decision-making. Traditional imitation learning, relying on observation-action pairs for expert demonstrations, fails to capture the expert's memory mechanisms used in decision-making. To capture memory processes as demonstrations, we introduce the concept of \textbf{memory dependency pairs} $(p, q)$ indicating that events at time $p$ are recalled for decision-making at time $q$. We introduce \textbf{AttentionTuner} to leverage memory dependency pairs in Transformers and find significant improvements across several tasks compared to standard Transformers when evaluated on Memory Gym and the Long-term Memory Benchmark. The code used in these experiments is available at \href{https://github.com/WilliamYue37/AttentionTuner}{https://github.com/WilliamYue37/AttentionTuner}.

\newpage

\begin{figure*}[ht]

\begin{center}
\centerline{\includegraphics[width=\textwidth]{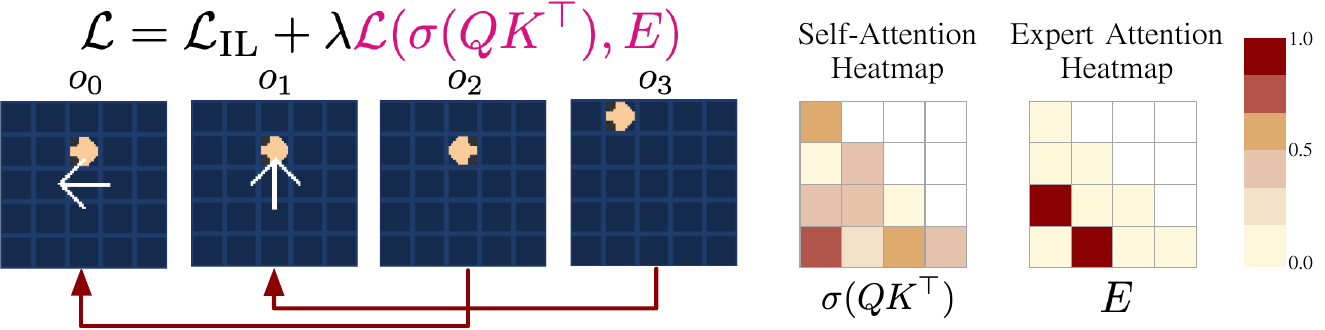}}
\caption{The red arrows indicate episodic memory dependencies labeled by an expert. The correct action to take in $o_2$ depends on $o_0$ and the correct action to take in $o_3$ depends on $o_1$. These memory dependency pairs are used to create the expert self-attention matrix $E \in \{0,1\}^{n \times n}$ where $n$ is the length of the sequence and $E_{ij} = 1$ only if the expert has indicated that observation $o_j$ should be recalled from memory at observation $o_i$, and $E_{ij} = 0$ otherwise. A binary cross entropy loss is taken between $E$ and the learner's self-attention matrix $\sigma(QK^\top)$ to form the memory loss $\mathcal{L}(\sigma(QK^\top), E)$ that encourages the learner to learn the expert's memory mechanism. The memory loss is scaled by $\lambda$ to match the magnitude of $\mathcal{L}_{\mathrm{IL}}$ and then added to form the final loss used during training.}
\label{fig1}
\end{center}

\end{figure*}

\section{Introduction}
\label{intro}

Partially Observable Markov Decision Processes (POMDPs) offer a framework for modeling decision-making in environments where the agent's information is incomplete, a common situation in real-world scenarios such as a robot operating based on limited camera observations. Making effective decisions under such conditions necessitates incorporating the agent's history, which can be encoded through memory mechanisms like Recurrent Neural Networks (RNNs) \cite{hausknecht2017deep, karkus2017qmdpnet, zhu2018improving, igl2018deep, hafner2019learning} or self-attention architectures such as Transformers \cite{esslinger2022deep}. 

However, it’s not always straightforward for a fully automated system to identify which points in history are crucial to remember for a particular decision. On the other hand, when humans learn, we are often taught not just the actions we need to take at the moment, but also which past events and memories should be recalled. For instance, a survival instructor might instruct students to recollect previously observed landmarks for navigating back to base camp or a coach could ask an athlete to recall a past encounter with an opponent when making their next move.

With this motivation in mind, this study concentrates on learning memory mechanisms essential for decision-making in POMDP tasks via expert demonstrations, also known as imitation learning. Standard imitation learning methods, which involve experts providing observation-action pairs, are insufficient in POMDP tasks as they do not capture the memory processes experts employ during decision-making.

To capture memory mechanisms through demonstration, we introduce the use of \textbf{memory dependency pairs} $(p, q)$, where $p < q$, indicating that the observation at time $p$ ought to be recalled for decision-making at time $q$. These memory dependency pairs can be integrated into the widely-used Transformers \cite{vaswani2023attention} by applying a loss to the self-attention matrix to reinforce attention between tokens representing times $p$ and $q$ (Figure~\ref{fig1}). 

The main contributions are as follows:
\begin{itemize}
    \item We introduce \textbf{memory dependency pairs} to incorporate memory mechanisms into demonstrations for imitation learning in POMDPs and to improve long-term credit assignment.
    \item We develop the \textbf{Long-term Memory Benchmark} (LTMB) for evaluating long-term memory capabilities in decision-making tasks.
    \item We introduce \textbf{AttentionTuner}, a novel method for leveraging memory dependency pairs in self-attention architectures, and benchmark it against vanilla Transformers on Memory Gym \cite{pleines2024memory} and LTMB. Empirical analyses show that AttentionTuner significantly improves success rates on four tasks and aids the optimizer in consistently navigating the loss landscape towards solutions with better generalizability compared to those found by optimizing the vanilla Transformer. Ablations reveal that these improvements in learning can be attained with as few as 0.1\% of demonstrations annotated.
\end{itemize}

\section{Related Work}
\label{related}

This section surveys the literature on memory types, integration of human feedback in decision-making algorithms, challenges in long-term credit assignment, and the development of memory mechanisms in learning models.

\subsection{Types of Memory}

Endel Tulving's 1972 research distinguishes between episodic memory, which stores information about specific events and their context, and semantic memory, a structured knowledge base for language and symbols \cite{Tulving1972, Tulving1983, tulving1985memory}. Long-term memory preserves a wide array of information, from skills to life events, over lengthy durations. In contrast, short-term or working memory, crucial for tasks like language comprehension and problem-solving, holds information briefly and allows for its manipulation, such as in mental arithmetic \cite{baddeley1992working, cowan2008differences}. While other dichotomies exist in the study of memory, such as declarative versus procedural \cite{humphreys1989different, ten1999procedural, ullman2004contributions}, and active versus inactive \cite{lewis1979psychobiology}, this work is primarily concerned with long-term episodic memory.

\subsection{Human Feedback in Decision Making}

Human feedback can be integrated into learning agents through various modalities. In reinforcement learning, scalar rewards may be assigned to individual states \cite{KCAP09-knox, SuttonBarto2018, warnell2018deep} or preferences may be expressed between trajectory pairs, either online or offline \cite{Wilson2012ABA, akrour2012april, wirth2016model, Sadigh2017ActivePL, lee2021pebble, stiennon2022learning, christiano2023deep}. In imitation learning, agents can learn from observation-action pairs, provided by humans in both online and offline contexts \cite{ross2011reduction, Zhang2017QueryEfficientIL, saunders2017trial, Torabi2018BehavioralCF}. Additional feedback mechanisms include gaze tracking \cite{ 8967843, saran2020efficiently}, binary corrective signals \cite{Celemin2019AnIF}, and human-provided trajectory outlines \cite{gu2023rttrajectory}. To the best of our knowledge, memory dependency pairs are the first modality through which humans can articulate their memory processes for decision-making.

\subsection{Long-Term Credit Assignment}
\label{credit}

The Long-Term Credit Assignment Problem highlights the difficulty of training agents to attribute consequences of actions over extended time frames \cite{sutton1984temporal, bengio1993credit, bengio1994learning}. Humans can base decisions on events from years or even decades past. However, agents today struggle with credit assignment over even short horizons. This challenge primarily arises from vanishing or exploding gradients during backpropagation through time or the memory mechanism's inability to retain relevant information amidst noise \cite{bengio1993credit}. Proposed solutions include adding skip connections to reduce backpropagation distances \cite{ke2018sparse, hung2019optimizing} and using self-attention in transformers \cite{vaswani2023attention}, which allows direct gradient flow between relevant timesteps. However, self-attention has been shown to not improve long-term credit assignment nor fully exploit all available information in its context \cite{liu2023lost, ni2023transformers}. In this work, memory dependency pairs are shown to assist self-attention in long-term credit assignment.

\subsection{Learning Memory Mechanisms}

RNNs were initially augmented with memory by incorporating hidden states and gating mechanisms, such as in Long Short-Term Memory (LSTM) networks \cite{Hochreiter1997LongSM, cho2014learning, burtsev2021memory}. Other approaches include integrating RNNs with differentiable memory that is key-addressable \cite{graves2014neural, weston2015memory, graves2016hybrid, wayne2018unsupervised}. Some researchers have also experimented with augmenting RNNs with stack-like memory modules \cite{joulin2015inferring}. Furthermore, combining LSTMs for short-term working memory with key-addressable memory for long-term episodic memory has been explored \cite{fortunato2020generalization}. Another significant development is the integration of Transformers with differentiable memory that can be either key or content addressable \cite{kang2023think, bessonov2023recurrent}. \citet{allen2024mitigating} uses the difference in temporal difference and Monte Carlo value estimates to detect partial observability and improve memory retention for RNNs. Our work is the first to explore learning memory mechanisms through expert demonstrations.

\section{Background}
\label{background}

This section defines the notation and framework for imitation learning in partially observable environments and provides a concise overview of Transformer architectures. This notation will be used to define AttentionTuner in Section~\ref{method}.

\subsection{Imitation Learning in Partially Observable Environments}
\label{IL_intro}

Imitation learning algorithms aim to learn a policy $\pi_\theta$ parameterized by $\theta$ by imitating a set of expert demonstrations $D = \{\tau_i\}_{i = 1 \ldots M}$. Each demonstration trajectory $\tau_i$ is a sequence of observation-action pairs ${(o_j, a_j)},~~j = 1 \ldots |\tau_i|$, where $|\tau_i|$ denotes the trajectory length. These trajectories are generated from a Partially Observable Markov Decision Process (POMDP), which is characterized by the tuple $\langle \mathcal{S}, \mathcal{O}, \mathcal{A}, T, O, \rho_0 \rangle$. Here, $\mathcal{S}$ represents the state space, $\mathcal{O}$ the observation space, $\mathcal{A}$ the action space, $T: \mathcal{S} \times \mathcal{A} \times \mathcal{S} \to [0, 1]$ the transition dynamics, $O: \mathcal{S} \times \mathcal{O} \to [0, 1]$ the observation function, and $\rho_0$ the initial state distribution. There are several approaches to imitation learning, including offline methods that do not require environmental interactions like behavioral cloning~\citep{schaal1999imitation}, online methods like Generative Adversarial Imitation Learning (GAIL) \cite{ho2016generative} and inverse reinforcement learning methods \cite{ng2000inverse}. In this study, we focus on behavioral cloning, where the objective is to minimize the negative log-likelihood loss function for a discrete action space:
\begin{equation}
\label{ILloss}
\mathcal{L}_{\text{IL}}(\theta) = -\mathbb{E}_{(o,y) \sim D}\bigg[\sum_a^A \mathds{1}_{y=a}\log(\pi(a \mid o))\bigg]
\end{equation}
where $A$ denotes the action space.

\subsection{Transformers for Decision Making}

\begin{figure}[h]
\begin{center}
\centerline{\includegraphics[width=\textwidth]{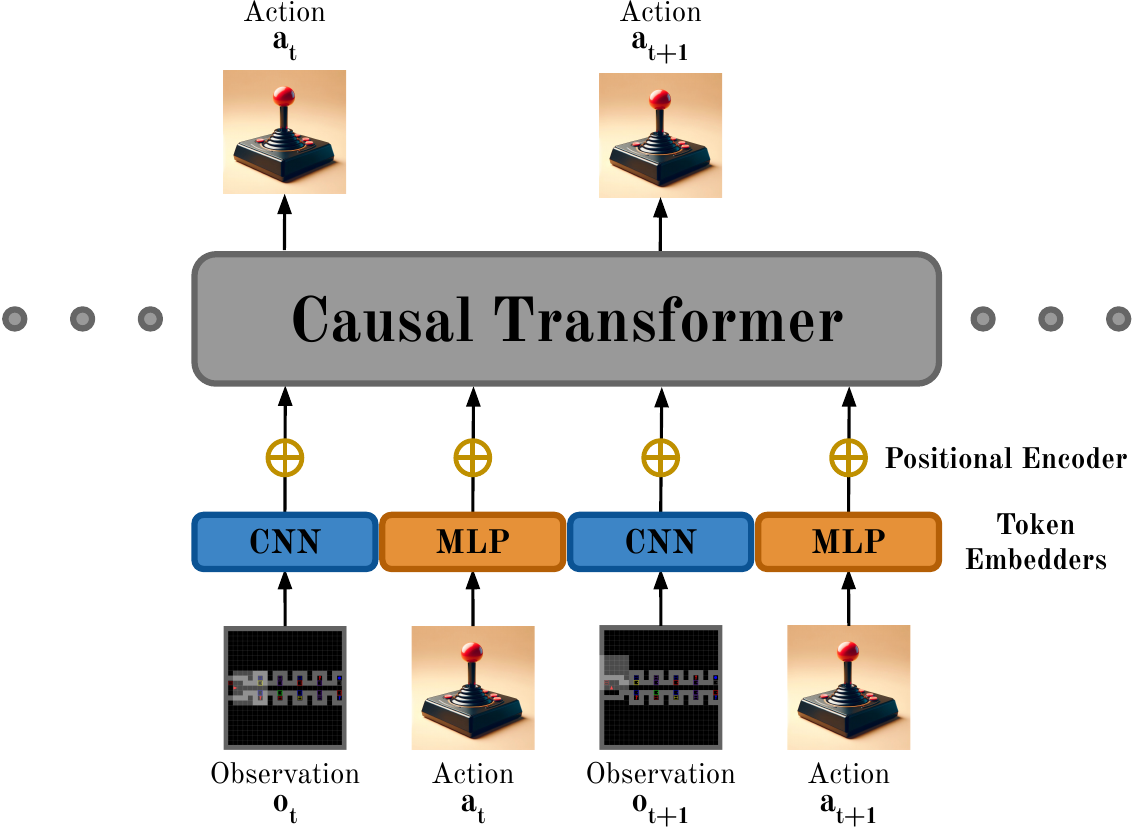}}
\caption{Architecture of the causal Transformer for sequential decision making modeling used in all AttentionTuner and vanilla Transformer experiments. The model embeds observations $o_t$ using a Convolutional Neural Network (CNN) and actions $a_t$ using a Multilayer Perceptron (MLP) network. Positional encodings are added to these embeddings to preserve positional context. Subsequently, they are fed into a causal Transformer decoder, which predicts future actions $a_{t}, a_{t+1}, \ldots$, conditioned on past events. Complete architectural details can be found in Appendix~\ref{architecture}.}
\label{transformer}
\end{center}
\end{figure}

Transformers are a neural network architecture designed to process sequential data \cite{vaswani2023attention}. The core of the Transformer architecture is the self-attention mechanism, which allows the model to weigh the importance of different parts of the input sequence. For any given sequence, the Transformer computes sets of queries $\mathbf{Q}$, keys $\mathbf{K}$, and values $\mathbf{V}$, typically through linear transformations of the input. Attention scores are calculated as follows:
\begin{equation}
    \text{Attention}(\mathbf{Q}, \mathbf{K}, \mathbf{V}) = \text{softmax}\left(\frac{\mathbf{QK}^T}{\sqrt{d_k}}\right)\mathbf{V}
\end{equation}
Here, \(d_k\) represents the dimension of the keys $\mathbf{K}$. To ensure causality in the Transformer model, masking is applied to the attention scores, particularly by zeroing out the upper right triangle of the attention matrix. This process ensures that a token is only influenced by preceding tokens. The self-attention mechanism equips the Transformer with an episodic memory, enabling it to emphasize previous tokens based on the similarity between query and key vectors.

The architecture extends this concept through multi-head self-attention, which involves computing attention scores multiple times with different sets of $\mathbf{Q}$, $\mathbf{K}$, and $\mathbf{V}$. Each Transformer layer consists of this multi-head self-attention mechanism, a feedforward network, and layer normalization. The complete Transformer decoder-only model is formed by stacking multiple such layers. Additionally, to retain positional information of tokens, positional encodings are added to the input embeddings before the first layer.

In the context of decision-making, trajectories can be modeled using the Transformer architecture, as illustrated in Figure~\ref{transformer}. This approach is analogous to the methods used in Decision Transformers \cite{Chen2021DecisionTR} and Trajectory Transformers \cite{Janner2021OfflineRL}. We represent a trajectory as:
\begin{equation}
    \tau = (o_1, a_1, o_2, a_2, \ldots, o_T, a_T)
\end{equation}
where $o_i$ and $a_i$ denote the observation and action, respectively, at timestep \(i\). For token embeddings, observations $o_i$ are processed through a convolutional neural network \cite{lecun1998gradient}, while actions $a_i$ are fed into a linear layer. The Transformer predicts the action $a_t$ for each observation $o_t$ as detailed in Figure~\ref{transformer}. These predictions are then utilized to compute the imitation learning training loss, as described in Equation~\ref{ILloss}.

Although Transformers excel in numerous domains, they encounter specific challenges in POMDP memory tasks. These challenges lead to the difficult and unstable optimization of the behavioral cloning objective (Equation~\ref{ILloss}), an issue further explored in Section~\ref{analysis}.

\section{Method}
\label{method}

We propose a novel approach to imitation learning by introducing \textbf{memory dependency pairs} within trajectories to address memory utilizations in decision-making. For each trajectory $\tau_i \in D$, we define $\mathcal{M}_i = \{(p_j, q_j)\}_{j = 1 \ldots |\mathcal{M}_i|}$ as the set of memory dependency pairs, where each pair $(p, q)$ indicates that observation $o_p$ was recalled during the decision-making process for $a_q$. If the decision-making process for $a_q$ depends on multiple past observations $o_{p_1}, \ldots, o_{p_n}$, then we can represent these dependencies with the pairs $(p_1, q), \ldots, (p_n, q)$. We extend the definition of an expert demonstration trajectory, initially described in Section~\ref{IL_intro}, to $\tau_i = \left(\{(o_j, a_j)\}_{j = 1 \ldots |\tau|}, \mathcal{M}_i\right)$, incorporating both observation-action pairs and memory dependencies. While in practice, a human would typically annotate memory dependency pairs (as shown in Appendix~\ref{sec:annotate}), in the experiments reported in this chapter, we instead use a computer program to automate the annotation of memory dependency pairs.

While in principle, memory dependency pairs could be used to enhance any memory-based learning architecture, in this thesis we introduce \textbf{AttentionTuner}, which specifically leverages them to enhance Transformer-based architectures. For each trajectory $\tau_i$, with length $n = |\tau_i|$, AttentionTuner constructs an expert self-attention matrix $E \in \{0,1\}^{n \times n}$, detailed in Figure~\ref{fig1}. This matrix, derived from $\mathcal{M}_i$, is defined as:
\[
E_{ij} = 
\begin{cases} 
1 & \text{if } (j, i) \in \mathcal{M}_i \\
0 & \text{otherwise}
\end{cases}
\]
To encourage the Transformer to mimic the expert's memory mechanism, AttentionTuner applies a binary cross-entropy loss between the expert matrix $E$ and the learner's self-attention matrix $A = \sigma(QK^\top) \in [0,1]^{n \times n}$. The memory loss equation is:
\begin{equation}
\label{memloss}
\begin{split}
    \mathcal{L}\left(A, E\right) &= -\frac{1}{n^2}\sum_{i = 1}^n\sum_{j = 1}^n\bigg[
    E_{ij}\log(A_{ij}) \\ &+ (1 - E_{ij})\log(1 - A_{ij})\bigg]
\end{split}
\end{equation}
In AttentionTuner, this memory loss is applied to a single attention head within the first Transformer layer (Figure~\ref{4x2}). The first layer is chosen because it is closest to the raw observation embeddings, making the application of memory dependency pairs more meaningful. Applying the loss to a single head allows other heads to learn additional memory mechanisms not captured by $\mathcal{M}_i$. Alternative applications of this memory loss are explored in Appendix~\ref{sec:mem-ablate}.

The memory loss is then scaled using a hyperparameter $\lambda$ and combined with the imitation learning loss $\mathcal{L_{\mathrm{IL}}}$ (defined in Equation~\ref{ILloss}) to form the final training loss:
\begin{equation}
    \mathcal{L} = \mathcal{L_{\mathrm{IL}}} + \lambda\mathcal{L}(A, E)
\end{equation}
We set $\lambda = 10$ based on robust performance observed across various benchmark tasks, effectively balancing the magnitude of the memory loss $\mathcal{L}(A, E)$ with the imitation learning loss $\mathcal{L_{\mathrm{IL}}}$. Comprehensive details of the model architecture, including the CNN and MLP embedders and the causal Transformer, are provided in Appendix~\ref{architecture}. Pseudocode for training AttentionTuner is provided in Algorithm~\ref{alg:attention_tuner}. The pseudocode for training vanilla Transformers is identical if the memory loss $\mathcal{L}_{\mathrm{memory}}$ is removed by setting $\lambda = 0$.

\begin{algorithm}[h]
\caption{AttentionTuner}
\label{alg:attention_tuner}
\begin{algorithmic}[1]
\Require Expert demonstrations $D = \{\tau_i\}_{i = 1 \ldots M}$, each $\tau_i = \left(\{(o_j, a_j)\}_{j = 1 \ldots |\tau|}, \mathcal{M}_i\right)$
\Require Hyperparameter for memory loss scaling $\lambda$
\Require Learning rate $\eta$
\Require Transformer model with CNN and MLP embedders, parameterized by $\theta$
\ForAll{epochs}
    \ForAll{$\tau_i \in D$}
        \State Extract observation-action pairs and memory dependency pairs: $\{(o_j, a_j)\}$, $\mathcal{M}_i$
        \State $o_{\mathrm{emb}} \leftarrow \mathrm{CNN}(\{o_j\})$
        \State $a_{\mathrm{emb}} \leftarrow \mathrm{MLP}(\{a_j\})$
        \State $\mathrm{input\_seq} \leftarrow \mathrm{PositionalEncoding}(o_{\mathrm{emb}}, a_{\mathrm{emb}})$
        \State $\{\hat{a}_j\}, A \leftarrow \mathrm{Transformer}(\mathrm{input\_seq})$ \Comment{$A$ is the self-attention matrix $\sigma(QK^\top)$ of the first attention head in the first Transformer layer}
        \State Initialize $E \leftarrow \{0\}^{|\tau_i| \times |\tau_i|}$
        \ForAll{$(p, q) \in \mathcal{M}_i$}
            \State $E[q][p] = 1$
        \EndFor
        \State $\mathcal{L}_{\mathrm{memory}} \leftarrow \mathrm{BinaryCrossEntropy}(A, E)$
        \State $\mathcal{L}_{\mathrm{IL}} \leftarrow \mathrm{NegativeLogLikelihood}(\{\hat{a}_j\}, \{a_j\})$
        \State $\mathcal{L} \leftarrow \mathcal{L}_{\mathrm{IL}} + \lambda \cdot \mathcal{L}_{\mathrm{memory}}$ 
        \State $\theta \leftarrow \theta - \eta\nabla_\theta\mathcal{L}$
    \EndFor
\EndFor
\end{algorithmic}
\end{algorithm}

\section{Experiments}

\begin{figure*}[ht]
\begin{center}
\centerline{\includegraphics[width=\textwidth]{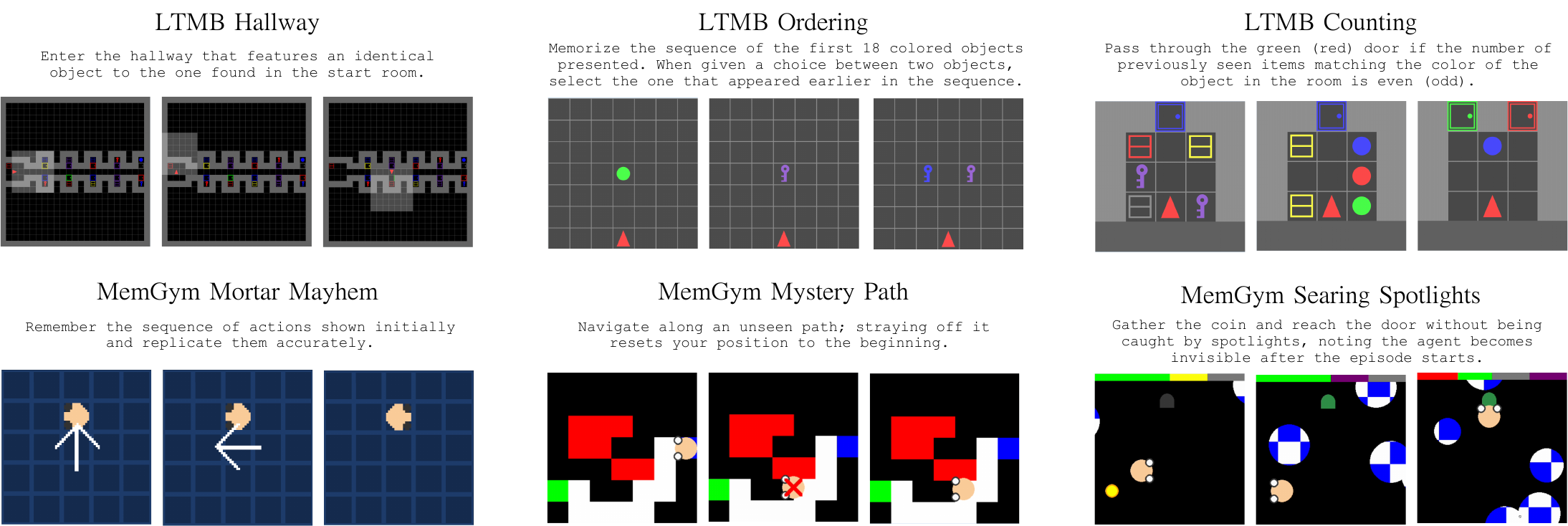}}
\caption{Overview of LTMB and MemGym tasks, each represented by a sequence of three images: the initial observation (left), a subsequent observation that must be recalled (middle), and a decision-making state that depends on memory of the middle state (right). Detailed task descriptions are available in Appendix~\ref{app:descrip}.}
\label{environments}
\end{center}

\end{figure*}

In this section, we conduct empirical assessments to 
\begin{enumerate}
    \item Determine whether AttentionTuner improves success rates and long-term credit assignment compared to the vanilla Transformer.

    \item Evaluate the value of memory dependency pair annotations for training in comparison to additional demonstrations.

    \item Assess the feasibility and human labor costs associated with annotating memory dependency pairs.
\end{enumerate}

\subsection{Experimental Setup}
We evaluated AttentionTuner on the Memory Gym benchmark \cite{pleines2023memory} and our newly introduced Long-term Memory Benchmark (LTMB), each featuring three procedurally generated POMDP tasks as illustrated in Figure~\ref{environments}. These tasks are specifically designed to necessitate and assess the use of memory mechanisms in agents. Four of the six tasks require memory dependencies on several past observations at a single timestep (shown in Appendix \ref{expert} Figure \ref{expert_heatmaps}). A key criterion for selecting these benchmarks was their simple environment dynamics, allowing a focus on episodic memory aspects within the tasks. Preference was given to gridworld environments whenever possible to simplify the process of hand-designing expert policies and automating memory dependency pair annotations. 

\paragraph{Baselines} Due to the lack of a good dense reward function in our benchmark environments and in most real world decision making tasks, we do not directly compare our method against methods that learn memory mechanisms through reinforcement learning, as they are expected to perform poorly on these tasks. We believe AttentionTuner focuses on a new problem setting, and we are not aware of other methods for learning memory mechanisms through demonstrations. For this reason, we compare our method against a vanilla Transformer as a baseline.

All experiments were conducted using at least 5 random seeds (Appendix~\ref{rand}), with each evaluated on 1,000 trials. Illustrations of sample expert attention matrices $E$ for all tasks are provided in Appendix~\ref{expert} Figure~\ref{expert_heatmaps}. Detailed environment settings and data collection for each task are provided in Appendix~\ref{settings}. Additionally, Appendix~\ref{hyperparameters} comprehensively documents the training hyperparameters utilized in our experiments. Finally, Appendix~\ref{compute} lists the computational resources allocated for these experiments.

\subsection{Results and Analysis}
\label{analysis}

\begin{table*}[ht]
\begin{center}
\resizebox{\textwidth}{!}{
\begin{small}
\begin{sc}
\begin{tabular}{lccccccr}
\toprule
    & Mortar & Mystery & Searing & & & \\
Methods & Mayhem & Path & Spotlights & Hallway & Ordering & Counting \\
\midrule
Vanilla Transformer & $20.8 \pm 42.2$ & $97.3 \pm 0.7$ & $62.2 \pm 5.5$ & $53.2 \pm 28.3$ & $59.4 \pm 22.9$ & $6 \pm 0.7$ \\
AttentionTuner (ours) & $\textbf{99.8} \pm 0.4$ & $\textbf{98.7} \pm 0.4$ & $\textbf{64.2} \pm 3.4$ & $\textbf{99.9} \pm 0.1$ & $\textbf{99.9} \pm 0.3$ & $\textbf{6.5} \pm 0.4$ \\
\bottomrule
\end{tabular}
\end{sc}
\end{small}
}
\end{center}
\caption{Average success rates and 90\% confidence intervals for two different methods—Vanilla Transformer and AttentionTuner (our approach)—across various tasks in the Memory Gym and Long-term Memory Benchmark.}
\label{benchmark}
\end{table*}

\begin{figure*}[ht]
\begin{center}
\centerline{\includegraphics[width=\textwidth]{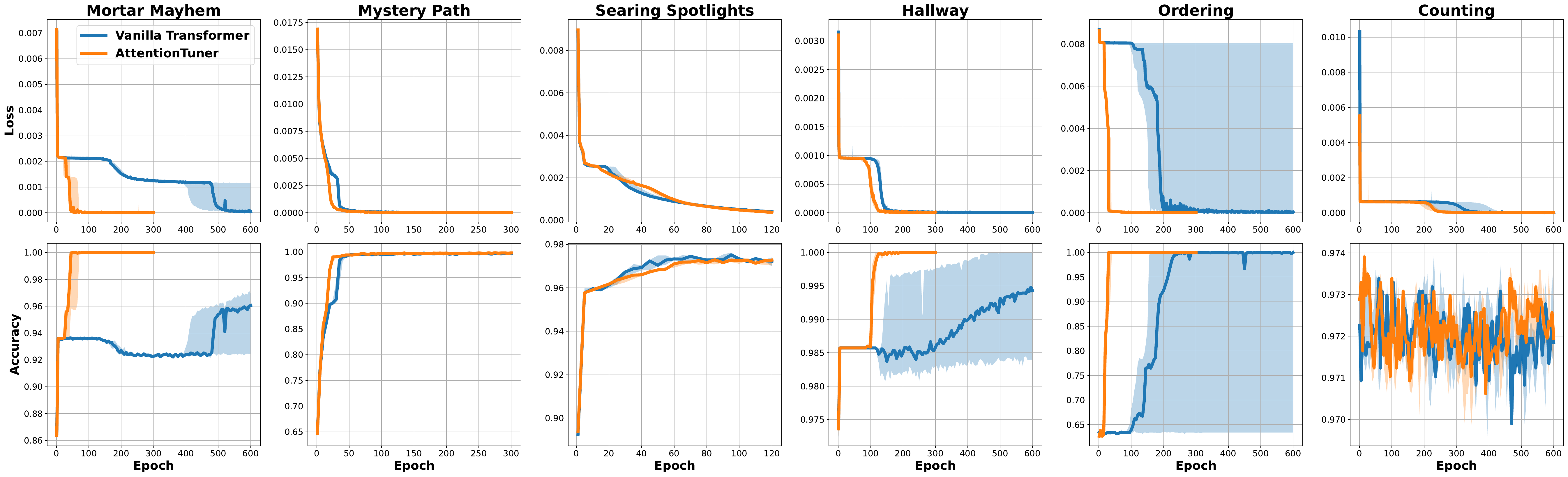}}
\caption{Median learning curves with interquartile range for Memory Gym and LTMB tasks are presented. The median is used rather than the mean due to the influence of outliers. Detailed mean learning curves are accessible in Appendix~\ref{mean_curves_section} Figure~\ref{mean_curves}. The top row displays the training loss while the bottom row shows the test accuracy on action prediction (not success rate). For a direct comparison with the vanilla Transformer, only the imitation learning loss (Equation~\ref{ILloss}) is plotted for AttentionTuner.} 
\label{med_curves}
\end{center}
\end{figure*}

\paragraph{AttentionTuner achieves significantly better success rates.}
In Table~\ref{benchmark}, performance of AttentionTuner is compared with the vanilla Transformer on Memory Gym and LTMB tasks. We performed a Welch t-test (results in Appendix~\ref{welch} Table~\ref{ttest}) with a significance threshold of $\alpha = 0.05$ and found that AttentionTuner achieves significantly better success rates on Mortar Mayhem, Mystery Path, Hallway, and Ordering. The high variability observed in the vanilla Transformer's performance on Mortar Mayhem, Hallway, and Ordering tasks can be attributed to some runs achieving nearly perfect success rates, while others fall close to zero.  This discrepancy is explained in the learning curves presented in Figure~\ref{med_curves}, indicating challenges in optimizing the behavioral cloning objective (Equation~\ref{ILloss}) for POMDP memory tasks with a large Transformer model. These challenges result in convergence to suboptimal local optima, as evidenced by flat segments in the training loss curves. 

\paragraph{AttentionTuner aids the optimizer in navigating the loss landscape.}
A local optimum is apparent in the Hallway and Counting tasks, while less pronounced in Mystery Path and Searing Spotlights (Figure~\ref{med_curves}). Notably, Mortar Mayhem and Ordering present two local optima, posing additional challenges for the optimizer. A unique observation in the Ordering task is that AttentionTuner encounters a single local optimum, in contrast to the vanilla Transformer, which faces two. The intricacies of task design that lead to the formation of multiple or difficult-to-escape local optima are not fully understood, highlighting an area for future research. 

The interquartile ranges in Figure~\ref{med_curves} suggest that AttentionTuner aids the optimizer in more efficiently and consistently escaping local optima, and potentially encountering fewer of them. For instance, in the Mortar Mayhem task, AttentionTuner enabled the optimizer to surpass both local optima in under 50 epochs with minimal variability across training seeds. In contrast, only a single training run of the vanilla Transformer overcame both local optima and did so with $\sim 400$ more training epochs. A similar pattern emerges in the Mystery Path, Hallway, Ordering, and Counting tasks where AttentionTuner consistently escapes from the local optima while the vanilla Transformer only escapes some of the time or escapes up to 200 epochs later than AttentionTuner (Figure~\ref{med_curves}). AttentionTuner's enhanced capability in traversing the loss landscape underscores its efficacy in facilitating long-term credit assignment and the learning of memory mechanisms.

\paragraph{AttentionTuner promotes convergence to solutions with better generalizabiltiy.}
Despite nearly identical learning curves after epoch 50 (Figure~\ref{med_curves}) on Mystery Path, AttentionTuner registers a significantly higher success rate than the vanilla Transformer. This difference in success rate suggests that even when the memory loss does not significantly improve training, it still promotes convergence to solutions with better generalizability. The differences in the final memory mechanisms learned by each model are illustrated in Appendix~\ref{learned_heatmaps}.

\paragraph{AttentionTuner exhibits limitations in learning short-term memory mechanisms.} In the Searing Spotlights task, AttentionTuner does not show a significant improvement in success rate or training efficiency. The expert attention matrix $E$ for this task (shown in Appendix~\ref{expert}), which attends to all previous actions is more resemblent of short-term memory than sparser long-term memory that our method is designed to optimize. Training an attention head to focus on all previous tokens effectively amounts to not focusing on any specific token, thereby diminishing the effectiveness of the memory loss. Despite efforts to distribute the memory loss across various attention heads, as investigated in Appendix \ref{sec:mem-ablate} Table~\ref{mem_ablation}, we did not observe any significant improvements. This outcome reinforces our understanding that the memory dynamics in the Searing Spotlights task are not ideally suited to the capabilities of AttentionTuner.

\subsection{Ablations}
\label{ablation}

In this section, we use ablations to assess the practicality of annotating memory dependency pairs. For each ablation study, experiments were run on the Mortar Mayhem and Hallway tasks with 5 random seeds.

\subsubsection{Imperfect Expert Annotations.}
\label{sec:annotate_ablate}

\begin{figure}[ht]
\begin{center}
\centerline{\includegraphics[width=\textwidth, keepaspectratio]{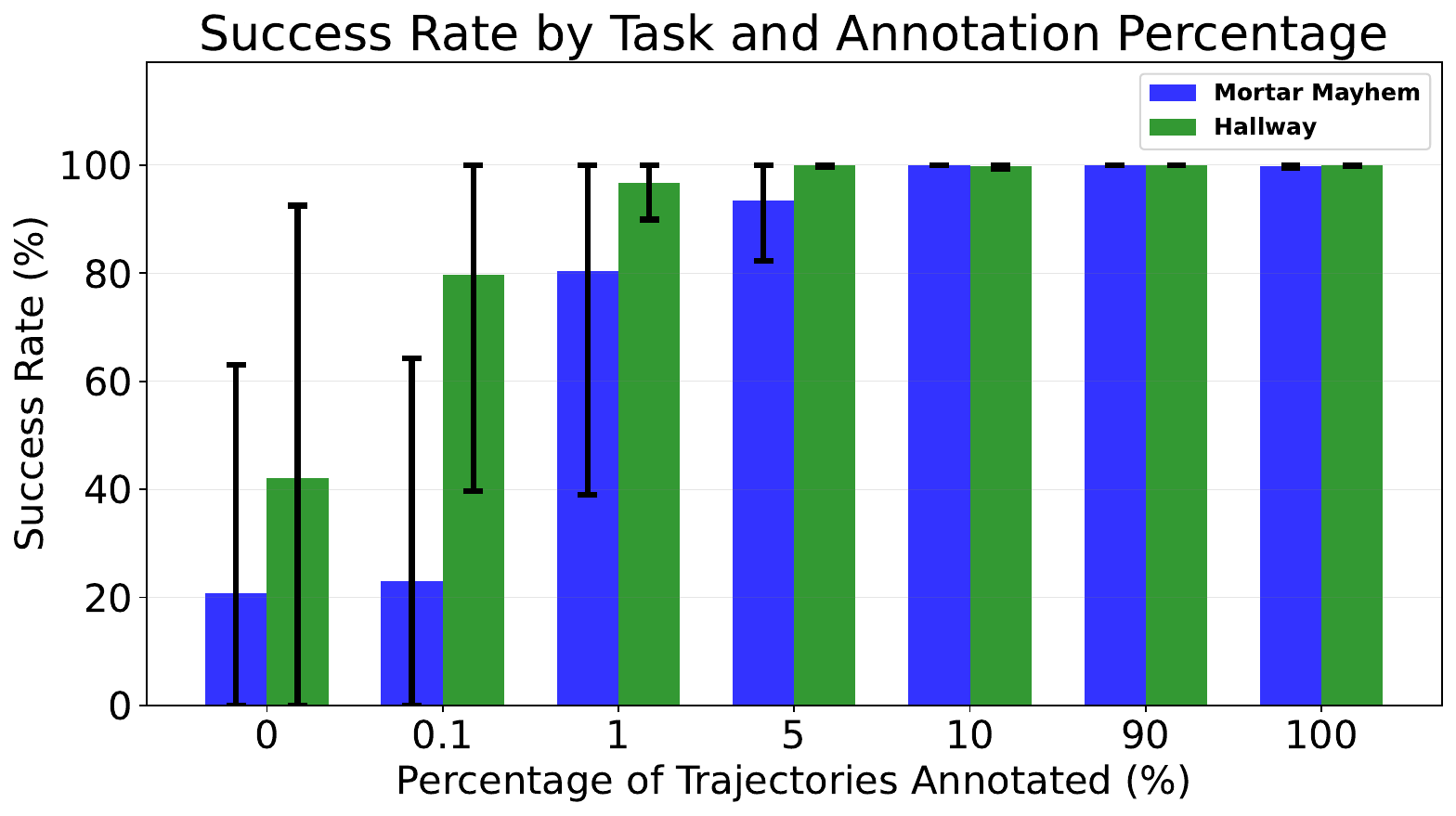}}
\caption{Success rates and 90\% confidence intervals for AttentionTuner training on Mortar Mayhem and Hallway tasks with missing annotations. Numerical values are presented in Appendix~\ref{sec:annotation} Table~\ref{tbl:annotations_ablation}.}
\label{annotations_ablation}
\end{center}

\end{figure}

Collecting comprehensive expert memory annotations for every demonstration trajectory in long complex decision-making tasks may not always be feasible. To explore the impact of this constraint, we conducted experiments varying the proportion of trajectories that included memory annotations, as detailed in Figure~\ref{annotations_ablation}. During training, the memory loss (Equation~\ref{memloss}) was only used on trajectories that included memory annotations. Trajectories without memory annotations were trained using only the imitation learning loss (Equation~\ref{ILloss}). The findings, presented in Figure~\ref{annotations_ablation}, reveal that having memory annotations for merely 10\% of the trajectories achieves comparable results to having annotations for all trajectories. Performance degradation for AttentionTuner becomes noticeable when the proportion of annotated demonstration trajectories falls below 10\%. Nonetheless, even with as few as 1\% of trajectories annotated, AttentionTuner manages to achieve a relatively high average success rate. Moreover, having even just 0.1\% of trajectories annotated still enables AttentionTuner to outperform the vanilla Transformer. 

Furthermore, human annotated trajectories could include errors. We found that AttentionTuner is robust against small perturbations of the memory dependency pair endpoints (Figure~\ref{tbl:imprecise_annotations_ablation},~\ref{tbl:imprecise_annotations_ablation2}). Potential solutions to make AttentionTuner more robust against larger perturbations are discussed in Section~\ref{sec:future}.

\begin{table}[h]
\caption{Partially Imprecise Annotations Ablations}
\label{tbl:imprecise_annotations_ablation}
\begin{center}
\resizebox{\textwidth}{!}{
\begin{small}
\begin{sc}
\begin{tabular}{lcccccc}
\toprule
Tasks & 0 & 1 & 2 & 5 & 10 & 20 \\
\midrule
Mortar Mayhem & $99.8 \pm 0.4$ & $99.22 \pm 0.7$ & $99.7 \pm 0.5$ & $100 \pm 0.1$ & $79.3 \pm 38.2$ & $21.2 \pm 42$ \\
Hallway & $99.9 \pm 0.1$ & $99.9 \pm 0.1$ & $82 \pm 38.5$ & $81.7 \pm 38.8$ & $88.4 \pm 23.7$ & $56.7 \pm 45.6$ \\
\bottomrule
\end{tabular}
\end{sc}
\end{small}
}
\end{center}
\caption*{The table presents success rates and their corresponding 90\% confidence intervals for tasks with varying levels of imprecise annotations. For each memory association pair $(p, q)$, the recalled timestep $p$ is perturbed by a delta $\Delta$ drawn from a normal distribution $\mathcal{N}(0, \sigma)$. The top row indicates the standard deviation $\sigma$.}

\end{table}

\begin{table}[h]
\caption{Imprecise Annotations Ablations}
\label{tbl:imprecise_annotations_ablation2}
\begin{center}
\resizebox{\textwidth}{!}{
\begin{small}
\begin{sc}
\begin{tabular}{lcccccc}
\toprule
Tasks & 0 & $0.5$ & $0.75$ & 1 & $1.5$ & 2 \\
\midrule
Mortar Mayhem & $99.8 \pm 0.4$ & $100 \pm 0$ & $99.4 \pm 1$ & $100 \pm 0$ & $57.5 \pm 50.2$ & $20 \pm 42.6$ \\
Hallway & $99.9 \pm 0.1$ & $100 \pm 0$ & $75.9 \pm 36.4$ & $20.3 \pm 24.2$ & $24.5 \pm 36$ & $3.5 \pm 2.6$ \\
\bottomrule
\end{tabular}
\end{sc}
\end{small}
}
\end{center}
\caption*{The table presents success rates and their corresponding 90\% confidence intervals for tasks with varying levels of imprecise annotations. For each memory association pair $(p, q)$, the recalled timestep $p$ and the timestep of recall $q$ are both perturbed by deltas $\Delta_p$ and $\Delta_q$ drawn from a normal distribution $\mathcal{N}(0, \sigma)$. The top row indicates the standard deviation $\sigma$.}
\end{table}

\subsubsection{Value of Memory Dependency Pairs Compared to Additional Demonstrations.}
\label{sec:value}

\begin{figure}[ht]
    \centering
    \includegraphics[width=\textwidth, keepaspectratio]{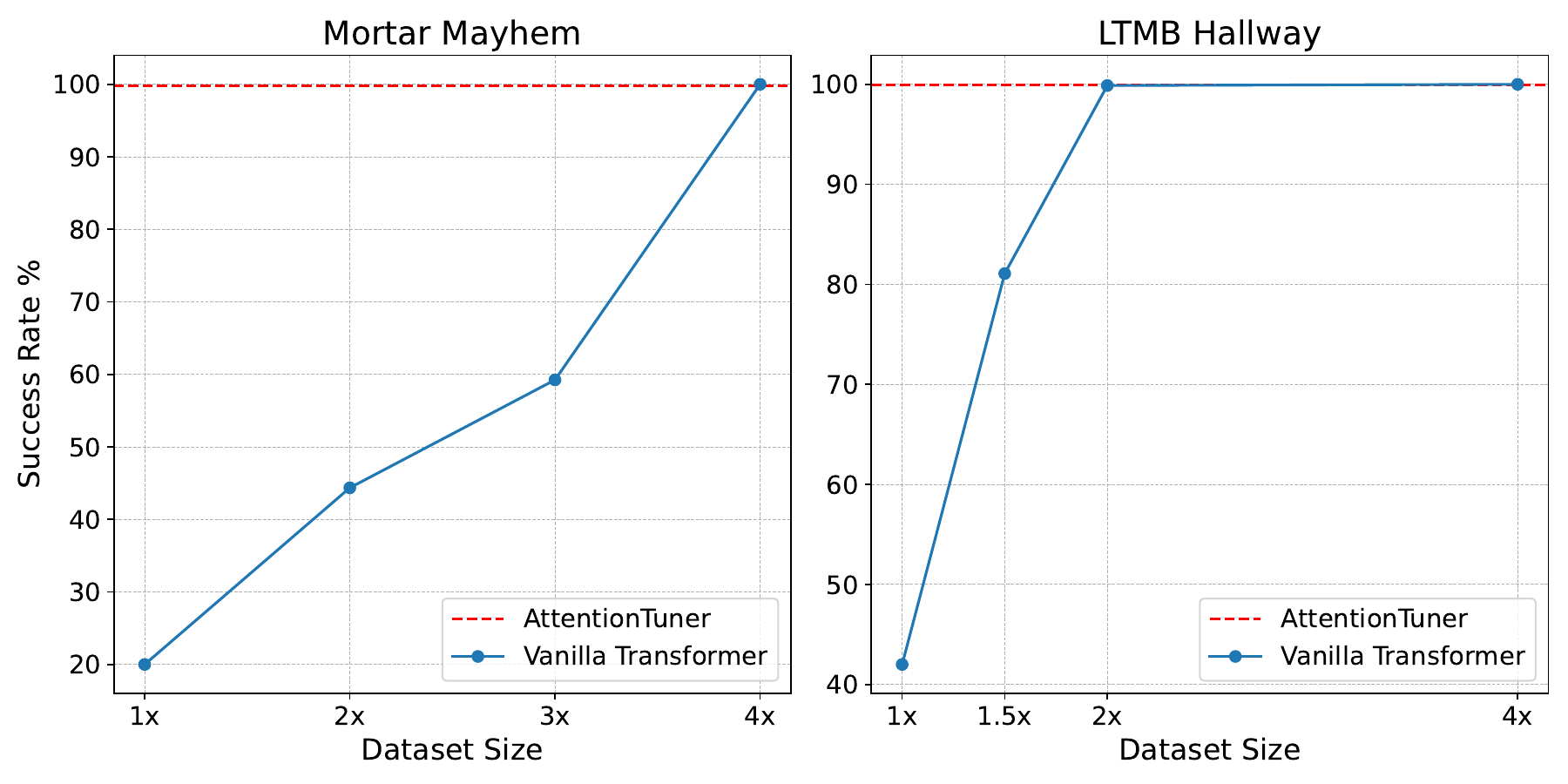}
    \caption{Success rates for vanilla Transformer with different training data sizes. The dotted red line represents AttentionTuner's success rate when training with a 1x dataset size. The number of demonstrations collected in the 1x datasets can be found in Appendix~\ref{settings}.}
    \label{fig:scaling}
\end{figure}

Figure~\ref{fig:scaling} shows that it can take anywhere from 2 to 4 times more demonstrations to train a vanilla Transformer to achieve the same success rate as AttentionTuner. In our experiments, we found that it takes 2 to 3 times longer for a human to annotate memory dependency pairs than to just collect a demonstration (Appendix~\ref{sec:annotate}). Considering that only 5-10\% of the demonstrations need to be annotated (Section~\ref{sec:annotate_ablate}), we find that memory dependency pair annotations on a single demonstration is equivalent to providing 40 additional demonstrations on both the Mortar Mayhem and Hallway tasks. Annotating memory dependency pairs instead of collecting additional demonstrations results in a human labor time savings factor of 16 on Mortar Mayhem and 14 on Hallway. Appendix~\ref{sec:math} details how these numbers are computed, though it is important to note that exact time savings can vary widely based on the task and the quality of the human annotator.

\chapter{Conclusion and Future Works}
\label{chp:conclusion}

In Chapters \ref{chp:lifelong} and \ref{chp:memory}, we introduced novel algorithms and frameworks for lifelong learning and memory learning from demonstrations, evaluating their effectiveness across various benchmarks. In this chapter, we summarize the key contributions of this thesis, discuss how this work advances the development of robots capable of operating over extended time horizons, and outline the future research needed to achieve the long-horizon capabilities required for large-scale robot deployment.

\newpage

\section{Summary}
We have introduced t-DGR, a novel method for continual learning in decision-making tasks, and introduced \textbf{memory dependency pairs} to enhance long-term credit assignment in POMDP tasks necessitating complex memory mechanisms.

t-DGR  has demonstrated state-of-the-art performance on the Continual World benchmarks. Our approach stands out due to its simplicity, scalability, and non-autoregressive nature, positioning it as a solid foundation for future research in this domain. Importantly, t-DGR takes into account essential properties of the real world, including bounded memory and blurry task boundaries. These considerations ensure that our method remains applicable and effective in real-world scenarios, enabling its potential integration into practical applications.

To utilize memory dependency pairs, we proposed AttentionTuner, a method designed for self-attention architectures to effectively leverage memory dependency pairs. Empirical evaluation on Memory Gym and LTMB benchmarks demonstrate that AttentionTuner effectively mitigates the local optima challenges in memory tasks, either by accelerating escape from such optima or by circumventing them entirely (Figure~\ref{med_curves}). This optimization improvement significantly increases success metrics across various tasks compared to vanilla Transformers (Table~\ref{benchmark}). Notably, these learning benefits are achievable with minimal annotation, requiring as few as $0.1\%$ of training trajectories to be annotated. This level of efficiency makes AttentionTuner a practical tool for real-world applications where learning complex memory mechanisms poses significant challenges.

Moreover, learning memory mechanisms and lifelong learning are not two separate isolated components needed for robots deployed at scale, but are interdependent capabilities. Learning long-horizon memory mechanisms necessitates agents capable of operating over extended lifespans and continuously improving their memory capabilities over that time span. Conversely, for an agent to learn effectively over its entire lifespan, it must have robust memory mechanisms to recall past mistakes and leverage previously acquired knowledge. Exploring methods to integrate t-DGR with AttentionTuner into a unified system presents an exciting direction for future research.

\section{Future Work}
\label{sec:future}

A potential avenue for future research in pseudo-rehearsal methods is the refinement of the replay mechanism employed in t-DGR. Rather than assigning equal weight to all past trajectories, a more selective approach could be explored such as generating more trajectories that the learner performs worse poorly on. The agent's learned memory mechanism could assist the generator in identifying relevant past experiences for replay. By prioritizing certain memories over others and strategically determining when to replay memories to the learner, akin to human learning processes, we could potentially enhance the performance and adaptability of our method. 

AttentionTuner primarily aims to enhance long-term episodic memory, thriving particularly when the expert attention matrix $E$ exhibits sparsity. However, it encounters limitations in scenarios involving short-term or semantic memory, a challenge exemplified by the performance in the Searing Spotlights task (Table~\ref{benchmark}). To mitigate this limitation, incorporating a short-term memory mechanism, like frame stacking, could be an effective strategy to complement AttentionTuner's long-term episodic memory.

The practicality of pinpointing precise timesteps for memory dependency pairs becomes cumbersome with longer episode horizons. To address this issue, a plausible direction could involve the summarization of token sequences into more generalized and abstract representations such as done in HCAM \cite{lampinen2021towards}. These summary ``token chunks" would allow for annotators to connect two events with natural language or approximate timesteps instead of having to connect two exact timesteps.

Another constraint stems from the Transformer model's mechanism of incorporating all preceding observations into its context, posing scalability challenges for tasks with extended horizons. Exploring hierarchical attention mechanisms or adopting a key-value cache system \cite{wu2022memorizing} presents promising avenues. Memory dependency pairs could serve as valuable assets in these contexts, guiding the prioritization and retention of pivotal events within each hierarchical layer or assisting in the optimization of key-value cache retrieval and management strategies.

While the focus of this work has been on integrating memory dependency pairs within the Transformer architecture, memory dependency pairs are applicable to a variety of neural architectures. For instance, in RNNs, a reconstruction loss on hidden states could promote memory retention, while in key-addressable, differentiable memory systems, a loss could encourage accurate key additions and queries. State space models can be viewed as minimizing an online learning objective \cite{liu2024longhornstatespacemodels}, and therefore memory dependency pairs can be used to emphasize which tokens the model should prioritize for retention (like a re-weighted regret). These ideas are left as an exciting frontier for future research endeavors.

% Bibliography

% Appendix
\newpage
\appendix
\chapter{Appendix}

\section{Hyperparameters}
\label{app:hyperparams}

\subsection{Finetune}

\begin{table}[H]
    \centering
    \label{tab:param:finetune}
    \resizebox{\textwidth}{!}{\begin{tabular}{|lcc|}
        \hline
        \textbf{Hyperparameter} & \textbf{Value} & \textbf{Brief Description} \\
        \hline
        batch size & 32 & number of samples in each training iteration \\
        epochs & 250 & number of times the entire dataset is passed through per task \\
        learning rate & $10^{-4}$ & learning rate for gradient descent \\
        optimization algorithm & Adam & optimization algorithm used \\
        $\beta_1$ & 0.9 & exponential decay rate for first moment estimates in Adam \\
        $\beta_2$ & 0.999 & exponential decay rate for second moment estimates in Adam \\
        epsilon & $10^{-8}$ & small constant for numerical stability \\
        weight decay & 0 & weight regularization \\
        \hline
    \end{tabular}}
\end{table}

\subsection{Multitask}

\begin{table}[H]
    \centering
    \label{tab:param:multitask}
    \resizebox{\textwidth}{!}{\begin{tabular}{|lcc|}
        \hline
        \textbf{Hyperparameter} & \textbf{Value} & \textbf{Brief Description} \\
        \hline
        batch size & 32 & number of samples in each training iteration \\
        epochs & 500 & number of times the entire dataset of all tasks is passed through \\
        learning rate & $10^{-4}$ & learning rate for gradient descent \\
        optimization algorithm & Adam & optimization algorithm used \\
        $\beta_1$ & 0.9 & exponential decay rate for first moment estimates in Adam \\
        $\beta_2$ & 0.999 & exponential decay rate for second moment estimates in Adam \\
        epsilon & $10^{-8}$ & small constant for numerical stability \\
        weight decay & 0 & weight regularization \\
        \hline
    \end{tabular}}
\end{table}

\subsection{oEWC}

\begin{table}[H]
    \centering
    \label{tab:param:owec}
    \resizebox{\textwidth}{!}{\begin{tabular}{|lcc|}
        \hline
        \textbf{Hyperparameter} & \textbf{Value} & \textbf{Brief Description} \\
        \hline
        batch size & 32 & number of samples in each training iteration \\
        epochs & 250 & number of times the entire dataset of all tasks is passed through \\
        learning rate & $10^{-4}$ & learning rate for gradient descent \\
        Fisher multiplier & $10^2$ & the Fisher is scaled by this number to form the EWC penalty \\
        optimization algorithm & Adam & optimization algorithm used \\
        $\beta_1$ & 0.9 & exponential decay rate for first moment estimates in Adam \\
        $\beta_2$ & 0.999 & exponential decay rate for second moment estimates in Adam \\
        epsilon & $10^{-8}$ & small constant for numerical stability \\
        weight decay & 0 & weight regularization \\
        \hline
    \end{tabular}}
\end{table}

The Fisher multiplier hyperparameter was tuned with the values:\\\(10^{-2}, 10^{-1}, 10^{0}, 10^1, 10^2, 10^3, 10^4, 10^5, 10^6\). We selected the value $10^2$ based on the success rate metric given by Equation~2. 

\subsection{PackNet}

\begin{table}[H]
    \centering
    \label{tab:param:packnet}
    \resizebox{\textwidth}{!}{\begin{tabular}{|lcc|}
        \hline
        \textbf{Hyperparameter} & \textbf{Value} & \textbf{Brief Description} \\
        \hline
        batch size & 32 & number of samples in each training iteration \\
        epochs & 250 & number of times the entire dataset of all tasks is passed through \\
        retrain epochs & 125 & number of training epochs after pruning \\
        learning rate & $10^{-4}$ & learning rate for gradient descent \\
        prune percent & 0.75 & percent of free parameters pruned for future tasks \\
        optimization algorithm & Adam & optimization algorithm used \\
        $\beta_1$ & 0.9 & exponential decay rate for first moment estimates in Adam \\
        $\beta_2$ & 0.999 & exponential decay rate for second moment estimates in Adam \\
        epsilon & $10^{-8}$ & small constant for numerical stability \\
        weight decay & 0 & weight regularization \\
        \hline
    \end{tabular}}
\end{table}

The retrain epochs and prune percent hyperparameters were chosen following the approach in the original PackNet paper. After training the first task, bias layers are frozen.

\subsection{DGR}

\begin{table}[H]
    \centering
    \label{tab:param:dgr}
    \resizebox{\textwidth}{!}{\begin{tabular}{|lcc|}
        \hline
        \textbf{Hyperparameter} & \textbf{Value} & \textbf{Brief Description} \\
        \hline
        batch size & 32 & number of samples in each training iteration \\
        epochs & 250 & number of times the entire dataset of all tasks is passed through \\
        learning rate & $10^{-4}$ & learning rate for gradient descent \\
        diffusion training steps & $10^4$ & number of training steps for the diffusion model per task \\ 
        diffusion warmup steps & $5 * 10^4$ & number of extra training steps for the diffusion model on the first task \\
        diffusion timesteps & $10^3$ & number of timesteps in the diffusion process \\ 
        replay ratio & 0.9 & percentage of training examples that are generated \\
        optimization algorithm & Adam & optimization algorithm used \\
        $\beta_1$ & 0.9 & exponential decay rate for first moment estimates in Adam \\
        $\beta_2$ & 0.999 & exponential decay rate for second moment estimates in Adam \\
        epsilon & $10^{-8}$ & small constant for numerical stability \\
        weight decay & 0 & weight regularization \\
        \hline
    \end{tabular}}
\end{table}

\subsection{CRIL}

\begin{table}[H]
    \centering
    \label{tab:param:cril}
    \resizebox{\textwidth}{!}{\begin{tabular}{|lcc|}
        \hline
        \textbf{Hyperparameter} & \textbf{Value} & \textbf{Brief Description} \\
        \hline
        batch size & 32 & number of samples in each training iteration \\
        epochs & 300 & number of times the entire dataset of all tasks is passed through \\
        learning rate & $10^{-4}$ & learning rate for gradient descent \\
        diffusion training steps & $10^4$ & number of training steps for the diffusion model per task \\ 
        diffusion warmup steps & $5 * 10^4$ & number of extra training steps for the diffusion model on the first task \\
        diffusion timesteps & $10^3$ & number of timesteps in the diffusion process \\ 
        replay ratio & 0.9 & percentage of training examples that are generated \\
        optimization algorithm & Adam & optimization algorithm used \\
        $\beta_1$ & 0.9 & exponential decay rate for first moment estimates in Adam \\
        $\beta_2$ & 0.999 & exponential decay rate for second moment estimates in Adam \\
        epsilon & $10^{-8}$ & small constant for numerical stability \\
        weight decay & 0 & weight regularization \\
        \hline
    \end{tabular}}
\end{table}

\subsection{t-DGR}

\begin{table}[H]
    \centering
    \label{tab:param:tdgr}
    \resizebox{\textwidth}{!}{\begin{tabular}{|lcc|}
        \hline
        \textbf{Hyperparameter} & \textbf{Value} & \textbf{Brief Description} \\
        \hline
        batch size & 32 & number of samples in each training iteration \\
        epochs & 300 & number of times the entire dataset of all tasks is passed through \\
        learning rate & $10^{-4}$ & learning rate for gradient descent \\
        diffusion training steps & $10^4$ & number of training steps for the diffusion model per task \\ 
        diffusion warmup steps & $5 * 10^4$ & number of extra training steps for the diffusion model on the first task \\
        diffusion timesteps & $10^3$ & number of timesteps in the diffusion process \\ 
        replay ratio & 0.9 & percentage of training examples that are generated \\
        optimization algorithm & Adam & optimization algorithm used \\
        $\beta_1$ & 0.9 & exponential decay rate for first moment estimates in Adam \\
        $\beta_2$ & 0.999 & exponential decay rate for second moment estimates in Adam \\
        epsilon & $10^{-8}$ & small constant for numerical stability \\
        weight decay & 0 & weight regularization \\
        \hline
    \end{tabular}}
\end{table}

\section{Model Architecture}
\label{app:model_arch}

\subsection{Multi-layer Perceptron}

\begin{table}[H]
    \centering
    \begin{tabular}{|l|l|l|}
        \hline
        \textbf{Layer (type)} & \textbf{Output Shape} & \textbf{Param \#} \\
        \hline
        Linear-1 & [32, 512] & 25,600 \\
        Linear-2 & [32, 512] & 262,656 \\
        Linear-3 & [32, 512] & 262,656 \\
        Linear-4 & [32, 512] & 262,656 \\
        Linear-5 & [32, 4]   & 2,052 \\
        \hline
        \multicolumn{2}{|r|}{Total params:} & 815,620 \\
        \multicolumn{2}{|r|}{Trainable params:} & 815,620 \\
        \multicolumn{2}{|r|}{Non-trainable params:} & 0 \\
        \hline
    \end{tabular}
    \caption{Multi-layer perceptron architecture of the learner shared by all methods, featuring five linear layers with ReLU activations applied after each layer except the final one.}
    \label{tab:model_architecture}
\end{table}

To condition the MLP learner on the task, the input vector is concatenated with a one-hot vector representing the task before being passed into the MLP.

\subsection{Diffusion U-net}

The U-net consists of 4 downsampling and 4 upsampling layers. To condition the generator U-net on the trajectory timestep, we use a sinusoidal positional encoder from Transformers to create an sinusoidal positional embedding that gets passed through 3 linear layers and then added to the convolutional layer output at each level of the U-net.

\subsection{Generative Adversarial Network}

The generator is implemented with an MLP consisting of 7 linear layers with batch normalization and Leaky ReLU activations applied after each layer except the final one. 

The discriminator is implemented with an MLP consisting of 3 linear layers with Leaky ReLU activations applied after each layer except the final one.

Sinusoidoal positional encodings are fed as input to a 2-layer MLP to condition both the generator and discriminator on the trajectory timestep for t-DGR.

\section{Experiment Details}
\label{app:experiment}

We utilized the following random seeds for the experiments: 1, 2, 3, 4, 5. All experiments were conducted on Nvidia A100 GPUs with 80 GB of memory. The computational node consisted of an Intel Xeon Gold 6342 2.80GHz CPU with 500 GB of memory. For our longest benchmark, CW20, the runtimes were as follows: DGR and t-DGR took 3 days, CRIL took 16 hours, finetune and oEWC took 6 hours, and PackNet took 8 hours.

\section{Neural Network Architectures}
\label{architecture}
Only the observation and action embedders differ in architecture between the two benchmarks. For both benchmarks, 4 Transformer layers were used with 2 self-attention heads per layer. Additionally, $d_{\text{model}} = 512$ for all experiments.  

\subsection{Causal Transformer}
\begin{figure}[H]
\begin{center}
\centerline{\includegraphics[width=\textwidth, height=0.8\textheight, keepaspectratio]{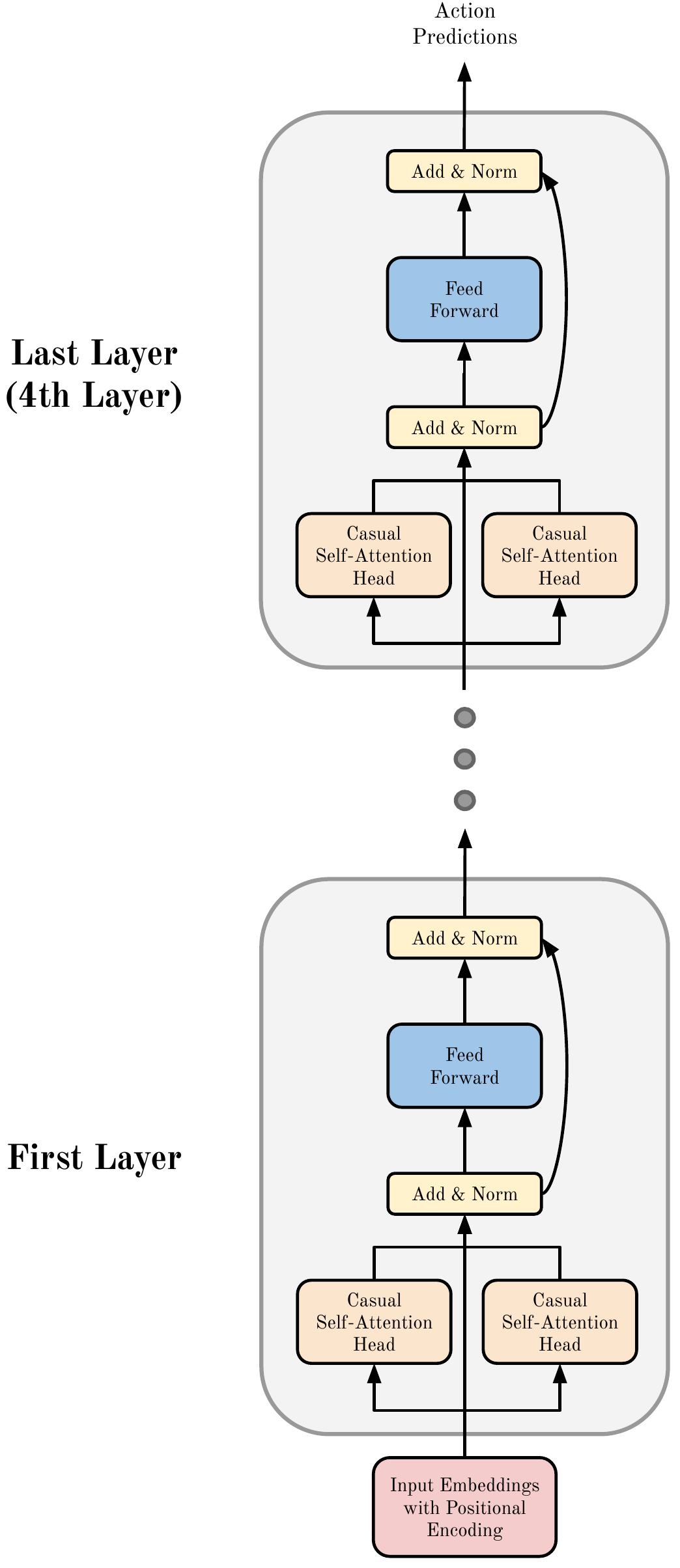}}
\caption{Causal Transformer architecture with 4 layers and 2 self-attention heads per layer as used in all experiments. The memory loss (Equation~\ref{memloss}) was applied to a single self-attention head in the first Transformer layer.}
\label{4x2}
\end{center}

\end{figure}

\subsection{Memory Gym}
\begin{lstlisting}
Transformer(
  (image_embedding): ImageEmbedding(
    (cnn): Sequential(
      (0): Conv2d(3, 32, kernel_size=(8, 8), stride=(4, 4))
      (1): ReLU()
      (2): Conv2d(32, 64, kernel_size=(4, 4), stride=(2, 2))
      (3): ReLU()
      (4): Conv2d(64, 64, kernel_size=(3, 3), stride=(1, 1))
      (5): ReLU()
    )
    (fc): Sequential(
      (0): Linear(in_features=3136, out_features=512, bias=True)
      (1): Tanh()
    )
  )
  (action_embedding): ActionEmbedding(
    (mlp): Sequential(
      (0): Linear(in_features=4, out_features=512, bias=True)
      (1): Tanh()
    )
  )
  (positional_encoding): PositionalEmbedding()
  (embedding_LN): LayerNorm((512,), eps=1e-05, elementwise_affine=True)
  (dropout): Dropout(p=0.1, inplace=False)
  (transformer_layers): ModuleList(
    (0-3): 4 x TransformerLayer(
      (self_attention): MultiheadAttention(
        (out_proj): NonDynamicallyQuantizableLinear(in_features=512, out_features=512, bias=True)
      )
      (norm1): LayerNorm((512,), eps=1e-05, elementwise_affine=True)
      (feedforward): Sequential(
        (0): Linear(in_features=512, out_features=2048, bias=True)
        (1): ReLU()
        (2): Linear(in_features=2048, out_features=512, bias=True)
        (3): Dropout(p=0.1, inplace=False)
      )
      (norm2): LayerNorm((512,), eps=1e-05, elementwise_affine=True)
    )
  )
  (output): Linear(in_features=512, out_features=4, bias=True)
)
\end{lstlisting}

\subsection{LTMB}
\begin{lstlisting}
Transformer(
  (image_embedding): ImageEmbedding(
    (cnn): Sequential(
      (0): Conv2d(20, 40, kernel_size=(3, 3), stride=(1, 1), padding=(1, 1))
      (1): ReLU()
      (2): Conv2d(40, 80, kernel_size=(3, 3), stride=(1, 1), padding=(1, 1))
      (3): ReLU()
    )
    (fc): Sequential(
      (0): Linear(in_features=3920, out_features=512, bias=True)
      (1): Tanh()
    )
  )
  (action_embedding): ActionEmbedding(
    (mlp): Sequential(
      (0): Linear(in_features=7, out_features=512, bias=True)
      (1): Tanh()
    )
  )
  (positional_encoding): PositionalEmbedding()
  (embedding_LN): LayerNorm((512,), eps=1e-05, elementwise_affine=True)
  (dropout): Dropout(p=0.1, inplace=False)
  (transformer_layers): ModuleList(
    (0-3): 4 x TransformerLayer(
      (self_attention): MultiheadAttention(
        (out_proj): NonDynamicallyQuantizableLinear(in_features=512, out_features=512, bias=True)
      )
      (norm1): LayerNorm((512,), eps=1e-05, elementwise_affine=True)
      (feedforward): Sequential(
        (0): Linear(in_features=512, out_features=2048, bias=True)
        (1): ReLU()
        (2): Linear(in_features=2048, out_features=512, bias=True)
        (3): Dropout(p=0.1, inplace=False)
      )
      (norm2): LayerNorm((512,), eps=1e-05, elementwise_affine=True)
    )
  )
  (output): Linear(in_features=512, out_features=7, bias=True)
)
\end{lstlisting}

\section{Application of the Memory Loss in Transformers}
\label{sec:mem-ablate}

\begin{table*}[ht]
\begin{center}
\resizebox{\textwidth}{!}{
\begin{small}
\begin{sc}
\begin{tabular}{lcccccccr}
\toprule
    & No Layer & First Layer & Middle Layer & Last Layer & First Layer & Middle Layer & Last Layer \\
Tasks & No Heads & All Heads & All Heads & All Heads & Single Heads & Single Heads & Single Heads \\
\midrule
Mortar Mayhem & $20.8 \pm 42.2$ & $\textbf{100} \pm 0.1$ & $99.7 \pm 0.6$ & $99.9 \pm 0.1$ & $99.8 \pm 0.4$ & $100 \pm 0.1$ & $39.9 \pm 52.1$ \\
Hallway & $42 \pm 50.5$ & $99.9 \pm 0.2$ & $\textbf{100} \pm 0$ & $89.4 \pm 22.5$ & $99.9 \pm 0.1$ & $99.8 \pm 0.3$ & $99.12 \pm 1.7$ \\
\bottomrule
\end{tabular}
\end{sc}
\end{small}
}
\end{center}
\caption{Success rate and 90\% confidence interval achieved on memory loss ablations}
\label{mem_ablation}
\end{table*}

The memory loss in Equation~\ref{memloss} has to be applied to a self-attention head in the Transformer. We posited in Section~\ref{method} that applying the memory loss to the first Transformer layer would make the memory dependency pairs more meaningful as the self-attention mechanism attends over the raw observation embeddings. This hypothesis is supported by the results in Table~\ref{mem_ablation}, which demonstrate that implementations applying the memory loss to the first layer consistently yield near-perfect success rates, surpassing other configurations. While we also speculated that dedicating memory loss to a single attention head would permit the remaining heads to engage in other memory processes, this distinction was not markedly evident in our results. The probable explanation is that memory dependency pairs in these relatively simple tasks sufficiently encapsulate all necessary memory functions, diminishing the benefit of isolating the memory loss to a single head. Attempting to distribute memory dependency pairs among various heads (Appendix~\ref{expert} Figure~\ref{split}), especially in the context of Searing Spotlights with its large amount of memory dependency pairs, did not yield a notable improvement in performance.

\section{Environment Descriptions}
\label{app:descrip}

\subsection{Memory Gym}
Memory Gym features three tasks—Mortar Mayhem, Mystery Path, and Searing Spotlights—set within a $7 \times 7$ gridworld. Agents receive $84 \times 84$ RGB image observations of the gridworld. For Mortar Mayhem and Mystery Path, the discrete action space includes: \texttt{move forward}, \texttt{turn left}, \texttt{turn right}, and \texttt{nop}. Searing Spotlights employs a multi-discrete action space, allowing movement in cardinal or ordinal directions plus a \texttt{nop} option.

\paragraph{Mortar Mayhem}
In this task, the agent memorizes and later executes a sequence of commands, indicated by arrows. An expert would annotate memory dependency pairs $(p, q)$, with $o_p$ representing the observation displaying the command and $o_q$ the observation of its execution.

\paragraph{Mystery Path}
Agents navigate a gridworld with an invisible path, restarting from the beginning if they deviate. To progress, they must remember their path to the deviation point. An expert would annotate memory dependency pairs $(p, q)$ where $o_p$ is a cell adjacent to the cell at $o_q$.

\paragraph{Searing Spotlights}
Agents aim to reach a door in a 2D plane after collecting a key, initially visible but obscured after 6 timesteps by dimming lights. The agent has to make it to the key and door while avoiding ``searing spotlights". An expert would annotate memory dependency pairs $(0, q), (1, q), \ldots, (q - 1, q)$ as agents must recall their starting position and all previous actions to deduce their current location. 

\subsection{Long-term Memory Benchmark}
The Long-term Memory Benchmark (LTMB) comprises three tasks: Hallway, Ordering, and Counting, set in the Minigrid environment \cite{MinigridMiniworld23}. Agents navigate a gridworld, receiving standard Minigrid $7 \times 7 \times 3$ state-based observations. The discrete action space includes \texttt{move forward}, \texttt{turn left}, \texttt{turn right}, and \texttt{nop}.

\paragraph{Hallway}
Agents identify and enter a hallway with an object matching one in the start room. The agent's view is limited to the $7 \times 7$ grid ahead. The expert annotates $(1, q)$ where timestep 1 represents the initial object observation.

\paragraph{Ordering}
Agents memorize the sequence of the first 18 colored objects encountered. When choosing between two objects, the agent selects the one appearing earlier in the sequence. An expert would annotate memory dependency pairs $(p, q)$, connecting the first observation $o_p$ of an object to the query observation $o_q$.

\paragraph{Counting}
In this task, agents traverse gallery rooms, memorizing six objects each, and query rooms, deciding which door to pass based on the parity of a query object's previous appearances. Passing through the wrong door ends the episode. An expert would annotate $(p, q)$ where $o_p$ is a gallery room containing the query object and $o_q$ is the query room.

\section{Environment Settings}
\label{settings}

Memory dependency pairs are annotated for every expert trajectory collected. In practice, memory dependency pairs would need to be annotated by the human demonstrator, for example via a graphical user interface of some sort. While it remains to be verified that this process can be made relatively seemless for human experts, for the purposes of this thesis, we sidestep this human-computer interaction issue by simulating both the expert demonstrations and the annotation of memory dependency pairs.

\subsection{Mortar Mayhem}
Discrete action movements were used with a command count of 10. Settings were default, except where noted. A total of 4,000 expert trajectories were collected, each with 118 timesteps.

\subsection{Mystery Path}
The origin and goal were not shown, with other settings at default. Training involved 4,000 expert trajectories, averaging 43 timesteps, and reaching up to 128 timesteps.

\subsection{Searing Spotlights}
The agent was visible for 6 timesteps before the lights dimmed completely. A single coin was used to unlock the exit. Other settings were left at the default. A single coin unlocked the exit, and other settings remained default. Training included 40,000 expert trajectories, averaging 30 timesteps, with a maximum of 75.

\subsection{Hallway}
The environment length was set to 30. A total of 5,000 expert trajectories were collected, averaging 67 timesteps, and maxing at 145 timesteps.

\subsection{Ordering}
The environment length was set to 50. A total of 5,000 expert trajectories were collected, each with a length of 68 timesteps.

\subsection{Counting}
The environment length was 20, with test room frequency at 30\% and empty tile frequency at 10\%. A total of 10,000 expert trajectories were collected, averaging 97 timesteps, with a maximum of 140.

\section{Training Hyperparameters}
\label{hyperparameters}

The following hyperparameters were shared by both AttentionTuner and the vanilla Transformer across all experiments:
\begin{table}[H]
\centering
\resizebox{\textwidth}{!}{\begin{tabular}{lll}
\toprule
\textbf{Hyperparameter} & \textbf{Value} & \textbf{Brief Description} \\
\midrule
batch size              & 64             & number of samples in each training iteration \\
learning rate           & $10^{-4}$      & learning rate for gradient descent \\
optimization algorithm  & Adam           & optimization algorithm used \\
$\beta_1$               & 0.9            & exponential decay rate for first moment estimates in Adam \\
$\beta_2$               & 0.999          & exponential decay rate for second moment estimates in Adam \\
epsilon                 & $10^{-8}$      & small constant for numerical stability \\
weight decay            & 0              & weight regularization \\
$\lambda$               & 10             & memory loss multiplier defined in Section~\ref{method} (only for AttentionTuner) \\
\bottomrule
\end{tabular}}
\caption{Hyperparameters used in experiments along with their brief descriptions}
\label{params}
\end{table}

Only the number of training epochs differed between methods and tasks.
\begin{table}[H]
\centering
\begin{tabular}{lcc}
\toprule
\textbf{Task}           & \textbf{AttentionTuner} & \textbf{Vanilla Transformer} \\
\midrule
Mortar Mayhem           & 300                     & 600                           \\
Mystery Path            & 300                     & 300                           \\
Searing Spotlights      & 120                     & 120                           \\
Hallway                 & 300                     & 600                           \\
Ordering                & 300                     & 600                           \\
Counting                & 600                     & 600                           \\
\bottomrule
\end{tabular}
\caption{Number of training epochs used for each task}
\label{epochs}
\end{table}

All experiments in Table~\ref{mem_ablation} (Section~\ref{sec:mem-ablate}) used 300 training epochs. All experiments in Figure~\ref{annotations_ablation} (Section~\ref{sec:annotate_ablate}) used 300 training epochs expect for vanilla Transformer runs (0\%) and Mortar Mayhem 0.1\% and 1\% which used 600 epochs.

\section{Random Seeds}
\label{rand}

All experiments were run with random seeds 1 through 5 except for the following:
\begin{itemize}
    \item 10 random seeds were used for vanilla Transformer on LTMB's Hallway task in Table~\ref{benchmark}.
    \item 15 random seeds were used for vanilla Transformer on LTMB's Ordering task in Table~\ref{benchmark}.
\end{itemize}
More training runs were used for these experiments due to their high variability.

\section{Welch t-test for Statistical Significance}
\label{welch}

We applied the Welch t-test to assess the statistical significance of the performance differences between AttentionTuner and the vanilla Transformer, as reported in Table~\ref{benchmark}.

\begin{table}[H]
\centering
\begin{tabular}{lc}
\toprule
\textbf{Task}           & \textbf{p-value} \\
\midrule
Mortar Mayhem           & 0.016 \\
Mystery Path            & 0.012 \\
Searing Spotlights      & 0.546 \\
Hallway                 & 0.014 \\
Ordering                & 0.008 \\
Counting                & 0.261 \\
\bottomrule
\end{tabular}
\caption{Welch t-test p-values for Performance Comparison Across Tasks}
\label{ttest}
\end{table}

\section{Average Learning Curves with 90\% Confidence Intervals}
\label{mean_curves_section}

\begin{figure}[H]
\begin{center}
\centerline{\includegraphics[width=\textwidth]{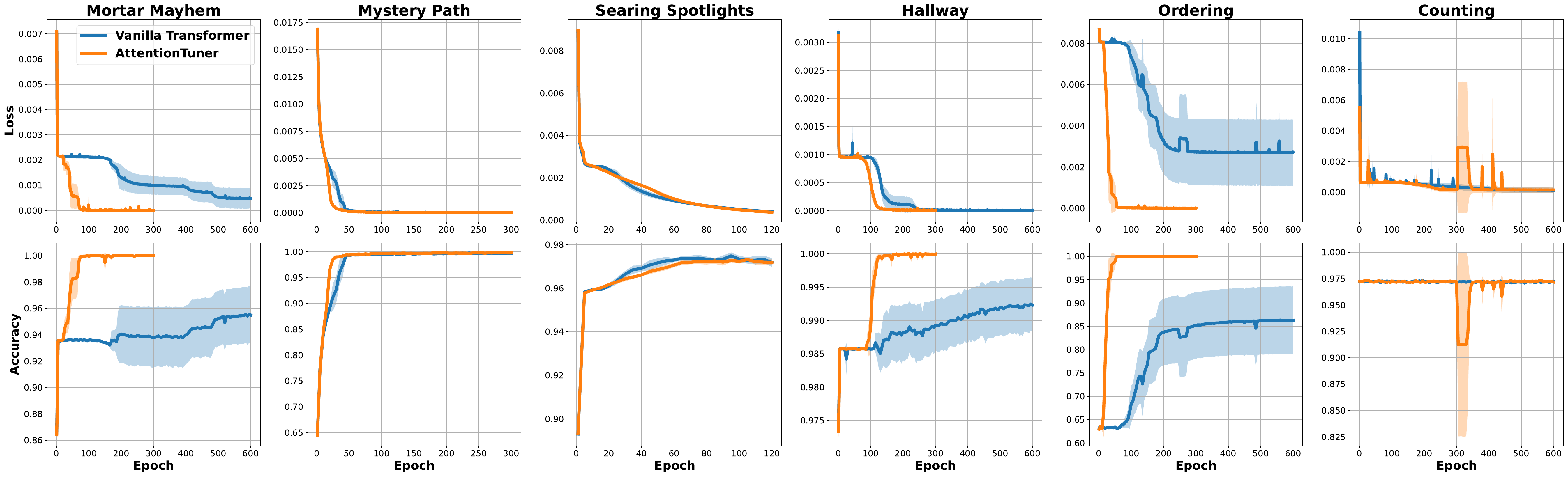}}
\caption{Mean learning curves with 90\% confidence intervals for Memory Gym tasks.}
\label{mean_curves}
\end{center}

\end{figure}

\section{Missing Annotations Ablation Data}
\label{sec:annotation}

\begin{table}[H]
\caption{Missing Annotations Ablations}
\label{tbl:annotations_ablation}
\begin{center}
\resizebox{\textwidth}{!}{
\begin{small}
\begin{sc}
\begin{tabular}{lcccccccr}
\toprule
Tasks & 0\% & 0.1\% & 1\% & 5\% & 10\% & 90\% & 100\% \\
\midrule
Mortar Mayhem & $20.8 \pm 42.2$ & $23 \pm 41.2$ & $80.3 \pm 41.3$ & $93.4 \pm 11.1$ & $100 \pm 0$ & $100 \pm 0$ & $99.8 \pm 0.4$ \\
Hallway & $42 \pm 50.5$ & $79.7 \pm 40$ & $96.7 \pm 6.8$ & $99.9 \pm 0.2$ & $99.8 \pm 0.5$ & $100 \pm 0$ & $99.9 \pm 0.1$ \\
\bottomrule
\end{tabular}
\end{sc}
\end{small}
}
\end{center}
\caption*{The table presents the success rates and their corresponding 90\% confidence intervals for tasks under different levels of missing annotations. The percentages indicate the proportion of demonstration trajectories with fully annotated memory dependency pairs. Figure~\ref{annotations_ablation} illustrates these results in a bar plot.}

\end{table}

\section{Annotating Memory Dependency Pairs}
\label{sec:annotate}

\begin{figure}[H]
\begin{center}
\centerline{\includegraphics[width=\textwidth, height=0.9\textheight, keepaspectratio]{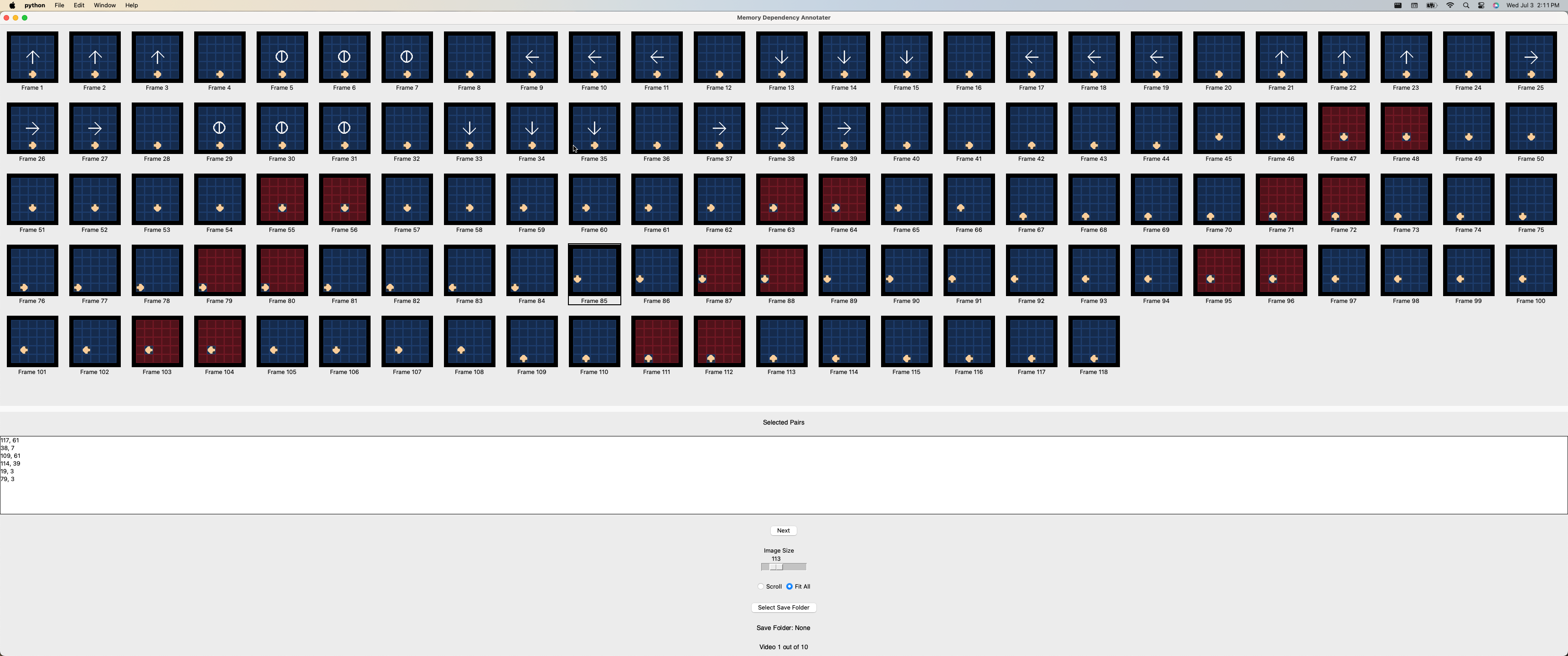}}
\caption{Graphical User Interface used to collect memory dependency pairs.}
\label{fig:gui}
\end{center}
\end{figure}

\begin{table}[H]
\centering
\begin{tabular}{lcc}
\toprule
\textbf{Task}           & \textbf{Demonstration Collection Time} & \textbf{Annotation Time} \\
\midrule
Mortar Mayhem           & 8 min 20 sec                     & 21 min 6 sec                           \\
Hallway                 & 2 min 9 sec                     & 5 min 59 sec                           \\
\bottomrule
\end{tabular}
\caption{The time it took to collect and annotate 10 demonstrations is recorded.}
\label{tbl:time}
\end{table}

Figure~\ref{fig:gui} shows the graphical user interface used to annotate demonstrations with memory dependency pairs. A single demonstration is loaded onto the GUI at a time. To annotate a memory dependency pair $(p, q)$, the annotator clicks on frame $p$ and then clicks on frame $q$. Table~\ref{tbl:time} compares the time it takes for humans to collect demonstrations versus annotating them.

\section{Computing the Value of Memory Dependency Pairs}
\label{sec:math}

\subsection{Mortar Mayhem}

There are 4000 demonstrations in the standard Mortar Mayhem training dataset (Appendix~\ref{settings}). Only 400 of these demonstrations have to be annotated to achieve a 100\% success rate (Figure~\ref{annotations_ablation}). It takes the vanilla Transformer 16,000 demonstrations to achieve a 100\% success rate (Figure~\ref{fig:scaling}). This means that each memory dependency pair annotation is worth $\frac{16000}{400} = 40$ additional demonstrations. According to Table~\ref{tbl:time}, annotating a single demonstration takes $\frac{1266}{10} = 126.6$ seconds while collecting 40 additional demonstration takes $\frac{500}{10} * 40 = 2000$ seconds. This results in a human labor time savings factor of $\frac{2000}{126.6} = 15.7977883 \approx 16$.

\subsection{Hallway}

There are 5000 demonstrations in the standard Hallway training dataset (Appendix~\ref{settings}). Only 250 of these demonstrations have to be annotated to achieve a 100\% success rate (Figure~\ref{annotations_ablation}). It takes the vanilla Transformer 10,000 demonstrations to achieve a 100\% success rate (Figure~\ref{fig:scaling}). This means that each memory dependency pair annotation is worth $\frac{10000}{250} = 40$ additional demonstrations. According to Table~\ref{tbl:time}, annotating a single demonstration takes $\frac{359}{10} = 35.9$ seconds while collecting 40 additional demonstrations takes $\frac{129}{10} *  40 = 516$ seconds. This results in a human labor time savings factor of $\frac{516}{35.9} = 14.3732591 \approx 14$.

\section{Expert Truth Attention Heatmaps}
\label{expert}

\begin{figure}[H]
\begin{center}
\centerline{\includegraphics[width=\textwidth, height=0.9\textheight, keepaspectratio]{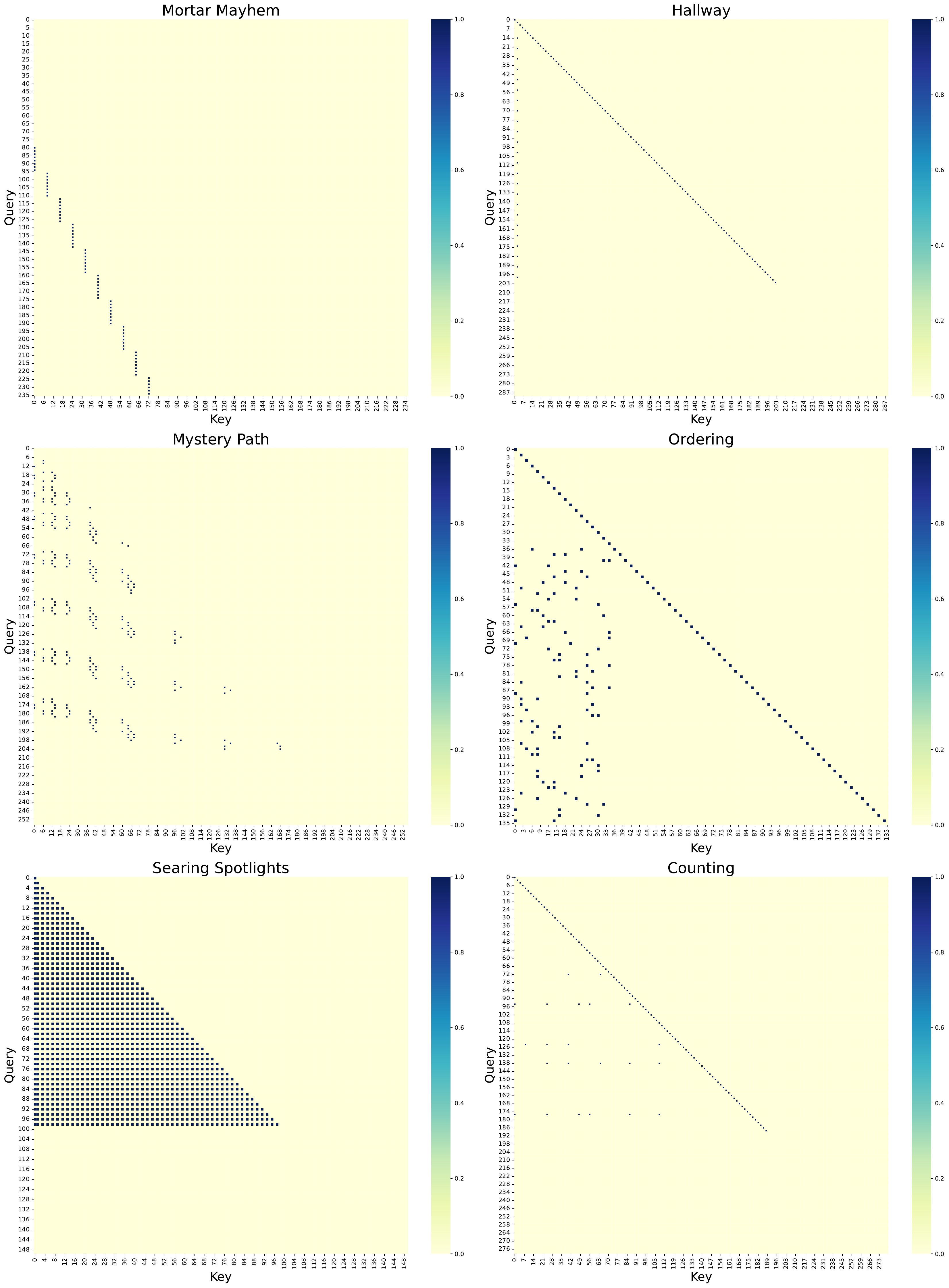}}
\caption{Sample expert attention heatmaps for $E$}
\label{expert_heatmaps}
\end{center}

\end{figure}

\begin{figure}[H]
\begin{center}
\centerline{\includegraphics[width=\textwidth]{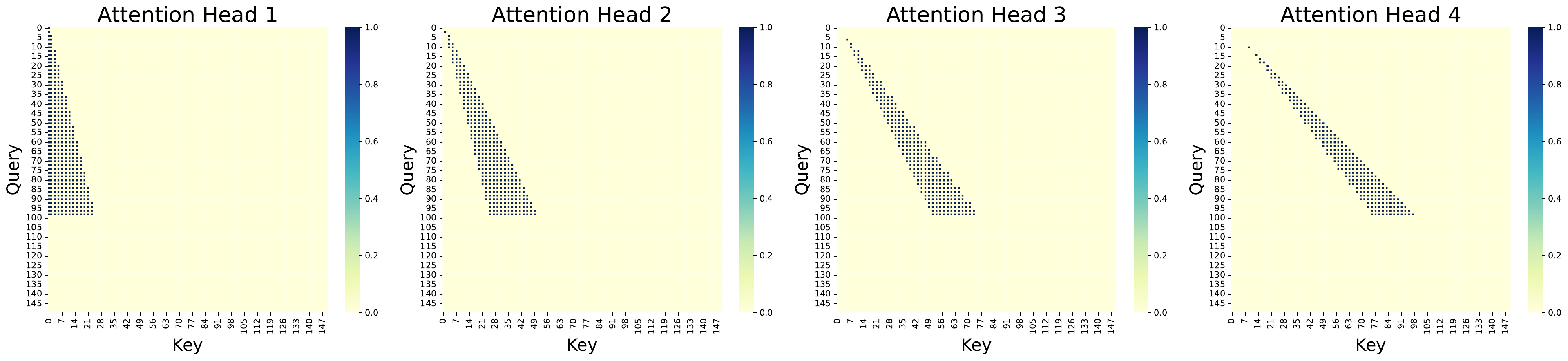}}
\caption{Sample expert attention heatmaps for split memory loss on Searing Spotlights. Notably, this experiment uniquely employs 4 self-attention heads, diverging from the typical configuration of 2 self-attention heads used in all other experiments.}
\label{split}
\end{center}

\end{figure}

\section{Learned Heatmaps}
\label{learned_heatmaps}

\begin{figure}[H]
\begin{center}
\centerline{\includegraphics[width=\textwidth, height=0.9\textheight, keepaspectratio]{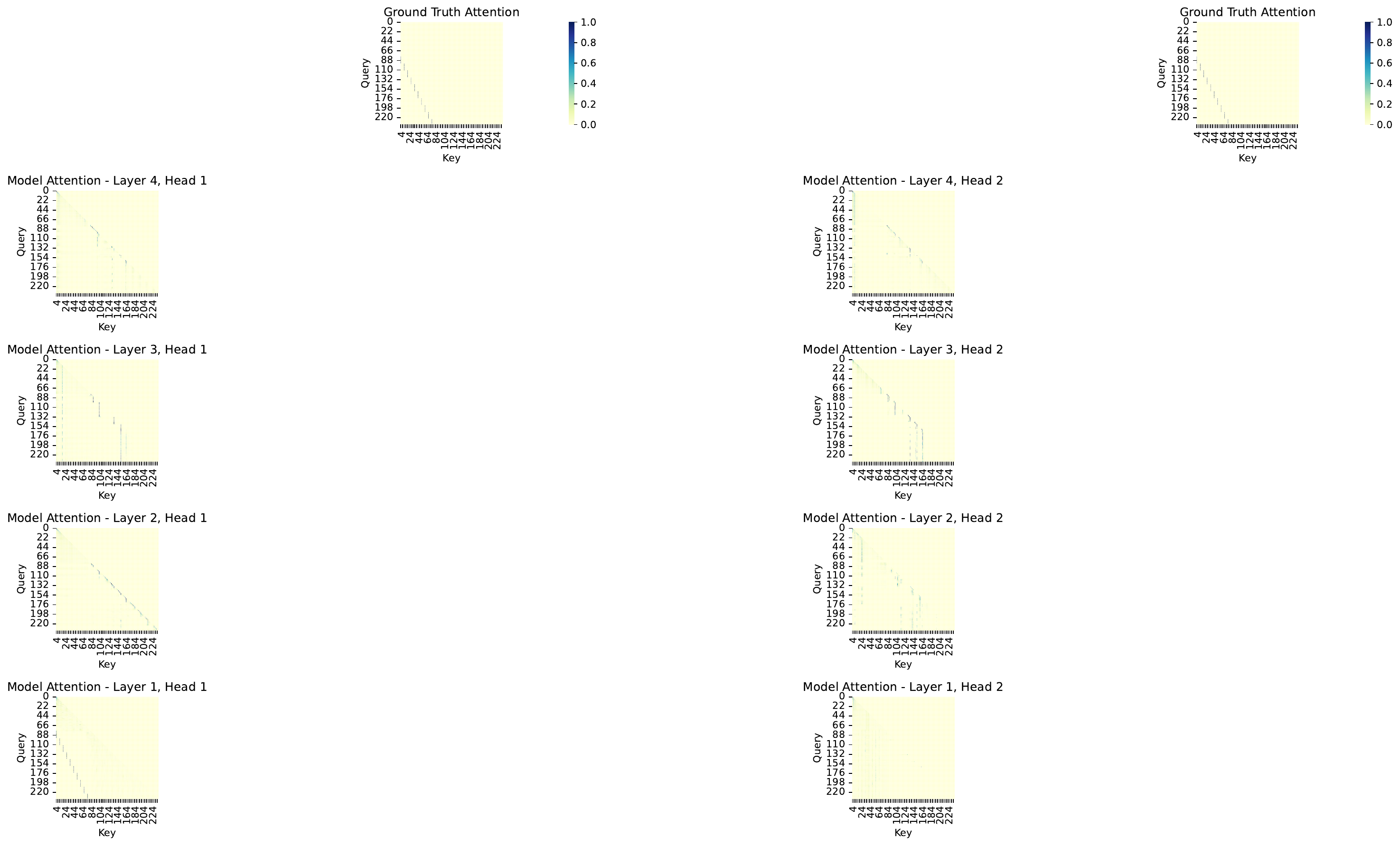}}
\caption{AttentionTuner's learned self-attention heatmap for all attention heads on Mortar Mayhem.}
\label{mem_heatmap}
\end{center}

\end{figure}

\begin{figure}[H]
\begin{center}
\centerline{\includegraphics[width=\textwidth, height=0.9\textheight, keepaspectratio]{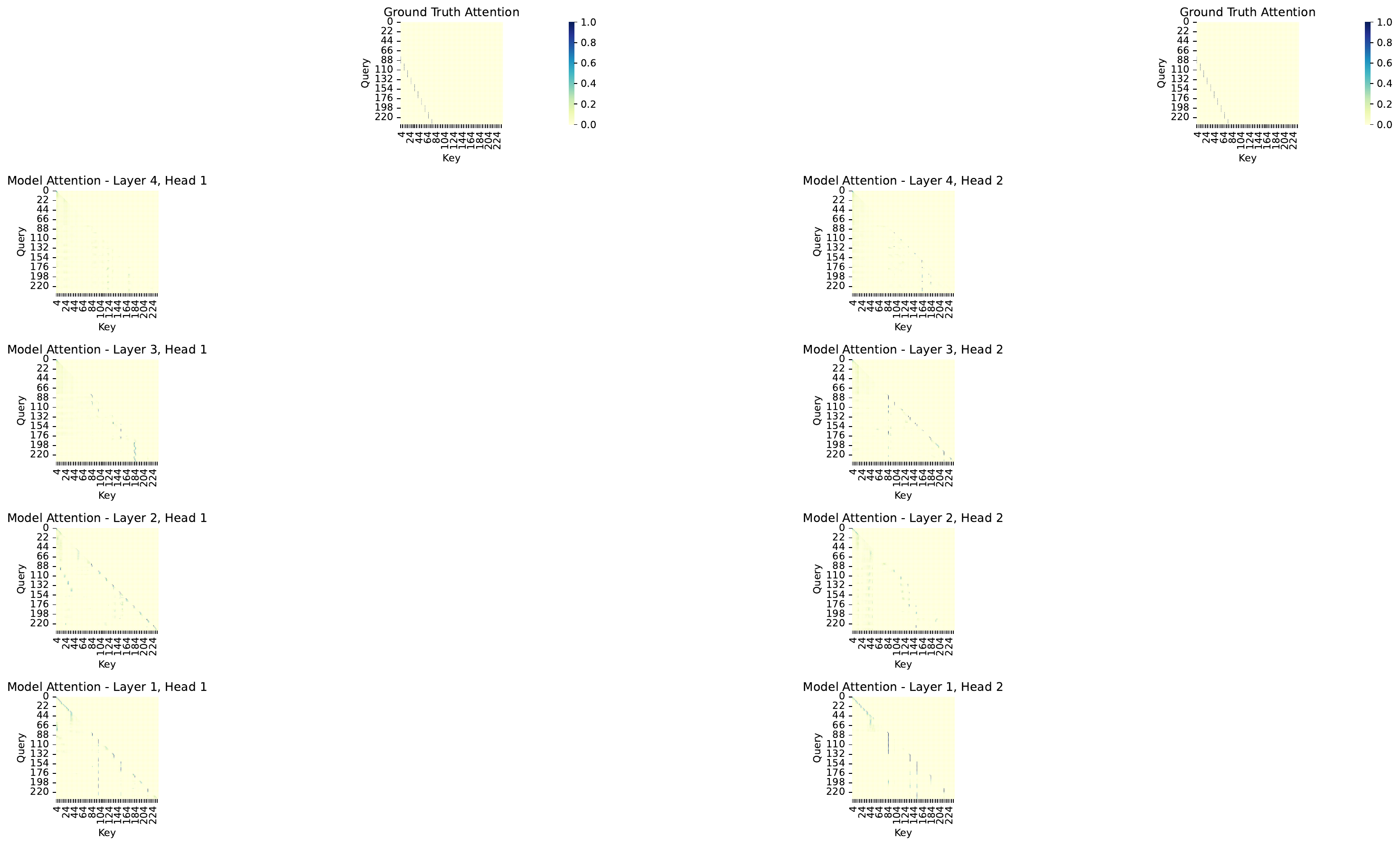}}
\caption{Vanilla Transformer's learned self-attention heatmap for all attention heads on Mortar Mayhem.}
\label{norm_heatmap}
\end{center}

\end{figure}

Figures \ref{mem_heatmap} and \ref{norm_heatmap} illustrate all self-attention heads in the Transformer as learned by AttentionTuner and the vanilla Transformer after training on Mortar Mayhem. Figure \ref{mem_heatmap} demonstrates that, with the application of memory loss (Equation~\ref{memloss}) to the initial head of the first layer, the correct memory mechanism was effectively acquired. Conversely, Figure \ref{norm_heatmap} indicates that in the absence of memory loss, solely employing the behavioral cloning objective did not facilitate the acquisition of the precise memory mechanism. However, a partially accurate memory mechanism is observable in the first head of the second layer.

\section{Computational Resources}
\label{compute}

All experiments were conducted on Nvidia A40 and A100 GPUs with 40 or 80 GB of memory. The computational node featured two Intel Xeon Gold 6342 2.80GHz CPUs with 500 GB RAM. Experiment durations varied between 1 and 4.5 hours.

\end{document}